\documentclass[10pt,journal,compsoc]{IEEEtran}

\usepackage{color}
\usepackage{epsfig}
\usepackage{graphicx}
\usepackage{amsmath}
\usepackage{amssymb}
\usepackage{multirow}
\usepackage{epstopdf}
\usepackage{subfigure}
\usepackage{url}
\usepackage{booktabs}
\usepackage{enumitem}
\usepackage{algorithm}
\usepackage{algpseudocode}

\ifCLASSOPTIONcompsoc
  \usepackage[nocompress]{cite}
\else
  \usepackage{cite}
\fi

\begin{document}

\title{TE141K: Artistic Text Benchmark for \\ Text Effect Transfer}

\author{Shuai~Yang$^*$,~\IEEEmembership{Student Member,~IEEE},
	    Wenjing Wang$^*$,~\IEEEmembership{Student Member,~IEEE},\\
        and Jiaying~Liu,~\IEEEmembership{Senior Member,~IEEE}
\thanks{$^*$ Equal contribution}
\thanks{This work is partially supported by National Natural Science Foundation of China under contract No.61772043, in part by Beijing Natural Science Foundation under contract No.L182002.
\textit{(Corresponding author: Jiaying Liu.)}}
\thanks{The authors are with Wangxuan Institute of Computer Technology, Peking University, Beijing 100871, China, (e-mail: \textnormal{williamyang@pku.edu.cn, daooshee@pku.edu.cn, liujiaying@pku.edu.cn}).}}

\IEEEtitleabstractindextext{
\begin{abstract}
  Text effects are combinations of visual elements such as outlines, colors and textures of text, which can dramatically improve its artistry. Although text effects are extensively utilized in the design industry, they are usually created by human experts due to their extreme complexity; this is laborious and not practical for normal users. In recent years, some efforts have been made toward automatic text effect transfer; however, the lack of data limits the capabilities of transfer models. To address this problem, we introduce a new text effects dataset, TE141K\footnote{Project page: \protect\url{https://daooshee.github.io/TE141K/}}, with 141,081 text effect/glyph pairs in total. Our dataset consists of 152 professionally designed text effects rendered on glyphs, including English letters, Chinese characters, and Arabic numerals. To the best of our knowledge, this is the largest dataset for text effect transfer to date. Based on this dataset, we propose a baseline approach called text effect transfer GAN (TET-GAN), which supports the transfer of all 152 styles in one model and can efficiently extend to new styles. Finally, we conduct a comprehensive comparison in which 14 style transfer models are benchmarked. Experimental results demonstrate the superiority of TET-GAN both qualitatively and quantitatively and indicate that our dataset is effective and challenging.
\end{abstract}

}

\maketitle

\IEEEdisplaynontitleabstractindextext
\IEEEpeerreviewmaketitle

\ifCLASSOPTIONcompsoc
\IEEEraisesectionheading{\section{Introduction}\label{sec:introduction}}
\else
\section{Introduction}
\fi

\IEEEPARstart{T}{EXT} effects are additional style features for text, such as colors, outlines, shadows, stereoscopic effects, glows and textures. Rendering text in the style specified by the example text effects is referred to as text effect transfer.
Applying visual effects to text is very common and important in graphic design.
However, manually rendering text effects is labor intensive and requires great skill beyond the abilities of normal users.
In this work, we introduce a large-scale text effect dataset to benchmark existing style transfer models on automatic text effect rendering and further propose a novel feature disentanglement neural network that can synthesize high-quality text effects on arbitrary glyphs.

Text effect transfer is a subtopic of general image style transfer.
General image style transfer has been extensively studied in recent years. Based on style representation, it can be categorized into global and local methods. Global methods~\cite{gatys2016image,Dumoulin2016A,huang2017adain,Li2017Demystifying,li2017universal} represent styles as global statistics of image features and transfer styles by matching global feature distributions between the style image and the generated image. The most famous one is the pioneering neural style transfer~\cite{gatys2016image}, which exploits deep neural features and represents styles as Gram matrices~\cite{gatys2015texture}. However, the global representation of general styles does not apply to text effects. Text effects are highly structured along the glyph and cannot be simply characterized in terms of the mean, variance or other global statistics~\cite{Li2017Demystifying,Dumoulin2016A,huang2017adain,li2017universal} of the texture features. Instead, the text effects should be learned with the corresponding glyphs.

On the other hand, local methods~\cite{Li2016Combining,Chen2016Fast,Yang2017Awesome} represent styles as local patches, and style transfer is essentially texture rearrangement, which seems to be more suitable for text effects than global statistics. In fact, the recent work of~\cite{Yang2017Awesome}, which is the first study of text effect transfer, is a local method; textures are rearranged to correlated positions on text skeletons.
However, it is difficult for local methods to preserve global style consistency. In addition, the patch-matching procedure of these methods usually suffers from a high computational complexity.

To handle a particular style, researchers have explored modeling styles from data rather than using general statistics or patches, which refers to image-to-image translation~\cite{Isola2017Image}. Early attempts~\cite{Isola2017Image,Zhu2017Unpaired} trained generative adversarial networks (GANs) to map images from two domains; this technique is limited to only two styles. StarGAN~\cite{Choi2017StarGAN} employs one-hot vectors to handle multiple predefined styles but requires expensive data collection and retraining to handle new styles.
Despite these limitations, image-to-image translation methods have shown great success in generating vivid styles of building facades, street views, shoes, handbags, \textit{etc.}, from the corresponding datasets~\cite{Tylecek13,doersch2012what,yu2014fine,zhu2016generative}.
However, the style of text effects is less well explored in this area due to the lack of related datasets.

\begin{figure*}[t]
  \centering
  \includegraphics[width=\linewidth]{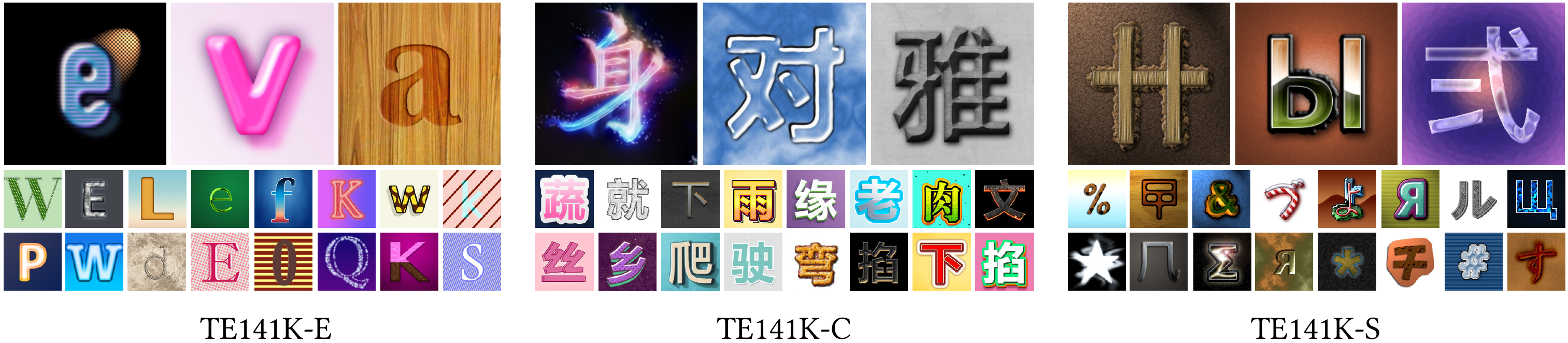}
  \caption{Representative text effects in TE141K. Text styles are grouped into three subsets based on glyph type, including TE141K-E~(English alphabet subset, 67 styles), TE141K-C~(Chinese character subset, 65 styles), and TE141K-S~(symbol and other language subset, 20 styles).}
  \label{fig:dataset}
\end{figure*}

An attempt to solve this problem uses MC-GAN~\cite{Azadi2017Multi},
for which a small-scale dataset containing $910$ images ($35$ text effects rendered on $26$ capital letters) is collected.
The authors~\cite{Azadi2017Multi} combined font transfer and text effect transfer using two successive subnetworks and trained them end-to-end using synthetic font data and the collected text effect dataset.
The data-driven model shows good performance on this dataset.
However, the limited size of the dataset can only support the model in handling capital letters with a small resolution of $64\times64$, which is far from meeting actual needs. Therefore, it is necessary to construct a large-scale dataset that has adequate style and glyph diversity for data-driven text effect transfer model design and benchmarking.

To address this practical issue, we develop TE141K\footnote{Project page: \protect\url{https://daooshee.github.io/TE141K/}}, a large-scale dataset with 141,081 text effect/glyph pairs for data-driven text effect transfer, as shown in Fig.~\ref{fig:dataset}.
TE141K contains 152 different kinds of professionally designed text effects collected from the Internet.
Each style is rendered on a variety of glyphs, including English letters, Chinese characters, Japanese kana, and Arabic numerals, to form the style images.
Besides these rendered in-the-wild styles, we design a simple yet effective style augmentation method~\cite{yang2019tet} to obtain infinite synthetic styles, which can serve as a supplement to TE141K to improve the robustness of transfer models.
Regarding content images, we further preprocess them so that they can provide more spatial information on glyphs, which makes it easier for the network to capture the spatial relationship between the glyph and the text effects.

\begin{figure}[t]
  \centering
   \includegraphics[width=0.93\linewidth]{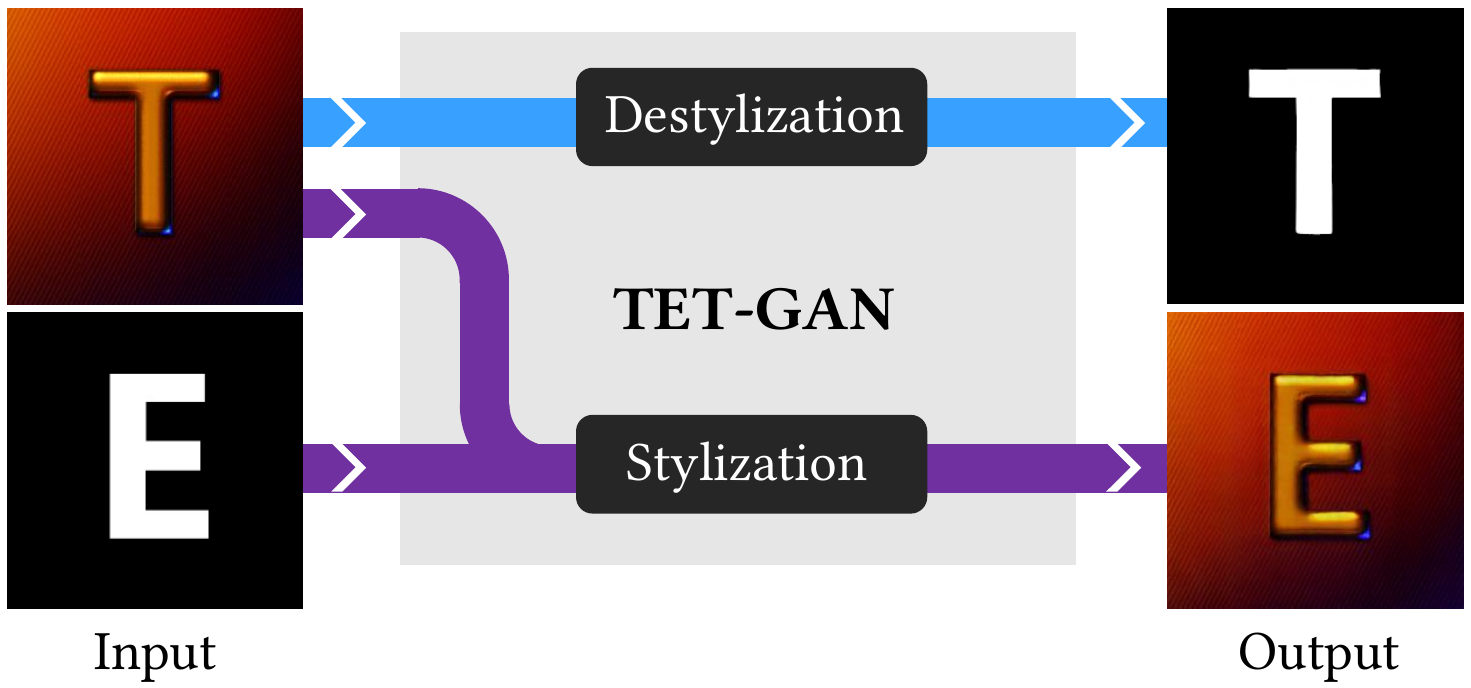}
  \caption{Our TET-GAN implements two functions: destylization for removing style features from the text and stylization for transferring the visual effects from highly stylized text onto other glyphs.}
  \label{fig:overview}
\end{figure}

Based on the large-scale dataset, we propose a novel approach for text effect transfer with two distinctive aspects.
First, we develop a novel TET-GAN built upon encoder-decoder architectures.
The encoders are trained to disentangle content and style features in the text effect images, while the decoders are trained to reconstruct features back to images.
TET-GAN performs two functions, stylization and destylization, as shown in Fig.~\ref{fig:overview}.
Stylization is implemented by recombining the disentangled content and style features, while destylization is implemented solely by decoding content features.
The task of destylization guides the network to precisely extract the content feature, which in turn helps the network better capture its spatial relationship with the style feature in the task of stylization.
Through feature disentanglement, our network can simultaneously support hundreds of distinct styles, whereas traditional image-to-image translation methods~\cite{Isola2017Image} only deal with two styles.
Second, we propose a self-stylization training scheme for one-reference style transfer. Leveraging the knowledge learned from our dataset, the network only needs to be finetuned on one reference example, and then it can render the new user-specified style on any glyph, providing much more flexibility than StarGAN~\cite{Choi2017StarGAN}.

\begin{table*}[t]
    \begin{center}
    \caption{A summary of TE141K. Based on glyph types, TE141K can be split into three subsets, where styles are different in different subsets.}
    \label{tb:dataset}
    \begin{tabular}{llllll}
    \toprule
    \textbf{} & \textbf{$\#$Style} & \textbf{$\#$Glyphs} & \textbf{Glyph Types} & \textbf{$\#$Training/$\#$Testing} & \textbf{$\#$Images}\\
    \midrule
    TE141K-E & 67 &  988 & 52 English Alphabets in 19 Fonts & 874/114 & 59,280 \\
    TE141K-C & 65 &  837 & 775 Chinese Characters, 52 English Alphabets, 10 Arabic Numerals & 740/97 & 54,405 \\
    TE141K-S & 20 &  1,024 & 56 Special Symbols, 968 Letters in Japanese, Russian, \textit{etc.} & 900/124 & 20,480 \\
    \midrule
    Total & 152 & 2,849 & & 2,514/335 & 141,081 \\
    \bottomrule
    \end{tabular}
    \end{center}
\end{table*}

Compared with our previous work~\cite{yang2019tet},
we expand TET-GAN to joint font style and text effect transfer, which achieves better style consistency between the output and the reference style.
We further explore semisupervised text effect transfer to improve the model's generalization.
In addition, we present analyses of the features extracted by TET-GAN to validate the disentanglement of styles and content, which helps clarify the working mechanism of the model.
Finally, in addition to our enclosed conference paper, we focus on the construction and analysis of the new large-scale dataset TE141K and conduct more comprehensive and in-depth experiments for model benchmarking, including 1) a new dataset $160\%$ larger than the old one~\cite{yang2019tet} in terms of styles and glyphs, 2) objective and subjective quantitative evaluations over fourteen style transfer models on three text effect transfer tasks, and 3) analyses on the performance-influencing factors of text effect transfer.
In summary, our contributions are threefold:
\begin{itemize}
\item We introduce a large dataset named TE141K containing thousands of professionally designed text effect images, which we believe can be useful for the research areas of text effect transfer, multidomain transfer, image-to-image translation, \textit{etc}.
\item We propose a novel TET-GAN to disentangle and recombine glyph features and style features for text effect transfer. The explicit content and style representations enable effective stylization and destylization on multiple text effects. A novel self-stylization training scheme for style extension is further proposed to improve the flexibility of the network.
\item We provide a comprehensive benchmark for our method and state-of-the-art methods, which validates the challenges of the proposed dataset and the superiority of our feature disentanglement model.
\end{itemize}

The rest of this paper is organized as follows.
Section~\ref{sec:dataset} presents our new dataset.
In Section~\ref{sec:related_work}, we review representative style transfer models for benchmarking.
In Section~\ref{sec:tetgan}, the details of the proposed TET-GAN for text effect transfer are presented.
Section~\ref{sec:experiment} benchmarks the proposed method and the state-of-the-art style transfer models.
Finally, we conclude our work in Section~\ref{sec:conclusion}.

\section{A Large-Scale Dataset for Text Effect Transfer}
\label{sec:dataset}

In this section, we will introduce the details of TE141K and analyze its data distribution.

\subsection{Data Collection}

We built a dataset with the help of automation tools in Adobe Photoshop.
Specifically, we first collected PSD files of text effects released by several text effect websites and PSD files we created by following tutorials on these websites.
Then, we used batch tools and scripts to automatically replace the glyphs and produce approximately one thousand text effect images for each PSD file.
There are also two text effects kindly provided by Yang~\textit{et al.}~\cite{Yang2017Awesome}, adding up to 152 different kinds of text styles.
Finally, we obtained 141,081 text effect images with a resolution of $320 \times 320$ and their corresponding glyph images to form TE141K.
Based on glyph types, we divide TE141K into three subsets, where text styles are different in different subsets. Fig.~\ref{fig:dataset} and Table~\ref{tb:dataset} show an overview of these three subsets, including:
\begin{itemize}
\setlength{\itemsep}{0pt}
\setlength{\parsep}{0pt}
\setlength{\parskip}{0pt}
\item \textbf{TE141K-E.} This subset contains 67 styles (59,280 image pairs, 988 glyphs per style), where all glyphs are English alphabets, making text effects easier to transfer compared to the other two subsets. This subset serves as a baseline to explore multistyle transfer.
\item \textbf{TE141K-C.} This subset contains 65 styles (54,405 image pairs, 837 glyphs per style). The glyphs for training are all Chinese characters, while the glyphs for testing contain Chinese characters, English alphabets and Arabic numerals. This subset can be used to test the glyph generalization ability of the transfer model.
\item \textbf{TE141K-S.} This subset contains 20 styles (20,480 image pairs, 1,024 glyphs per style). The glyphs are special symbols and letters of common languages other than Chinese and English. In this paper, we use this subset for one-reference training to test the flexibility (efficiency of new style extension) of the transfer model.
\end{itemize}
For each subset and each kind of text effect, we use approximately 87\% of the images for training and 13\% for testing.

\subsection{Data Processing}

\begin{figure}[t]
  \centering
  \includegraphics[width=\linewidth]{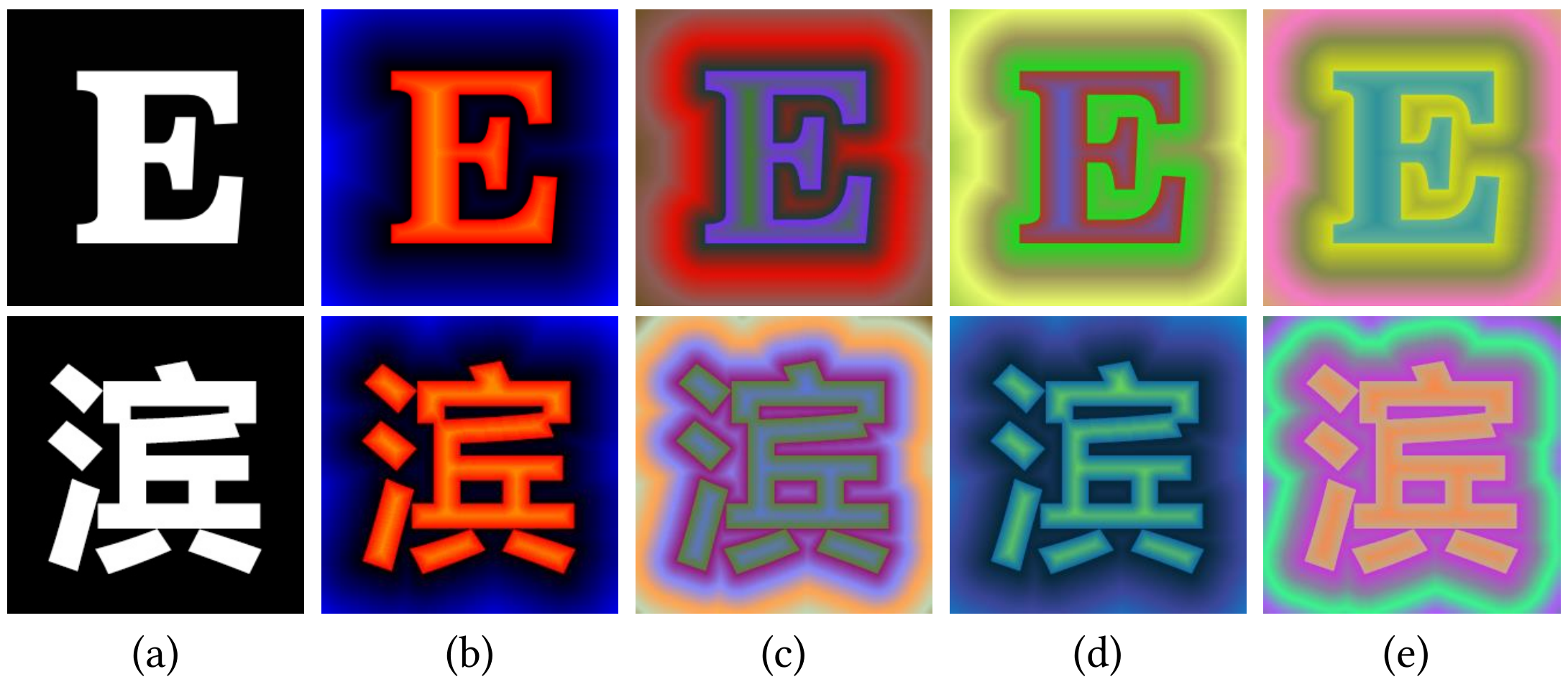}
  \caption{Distribution-aware data augmentation. (a) Raw text image.
  (b) Results of distribution-aware text image preprocessing (the contrast is enhanced for better visualization).
  (c)-(e) Results of distribution-aware text effect augmentation by tinting (b) using random colormaps.}\label{fig:preprocessing}
\end{figure}

\begin{figure}[t]
  \centering
  \includegraphics[width=0.98\linewidth]{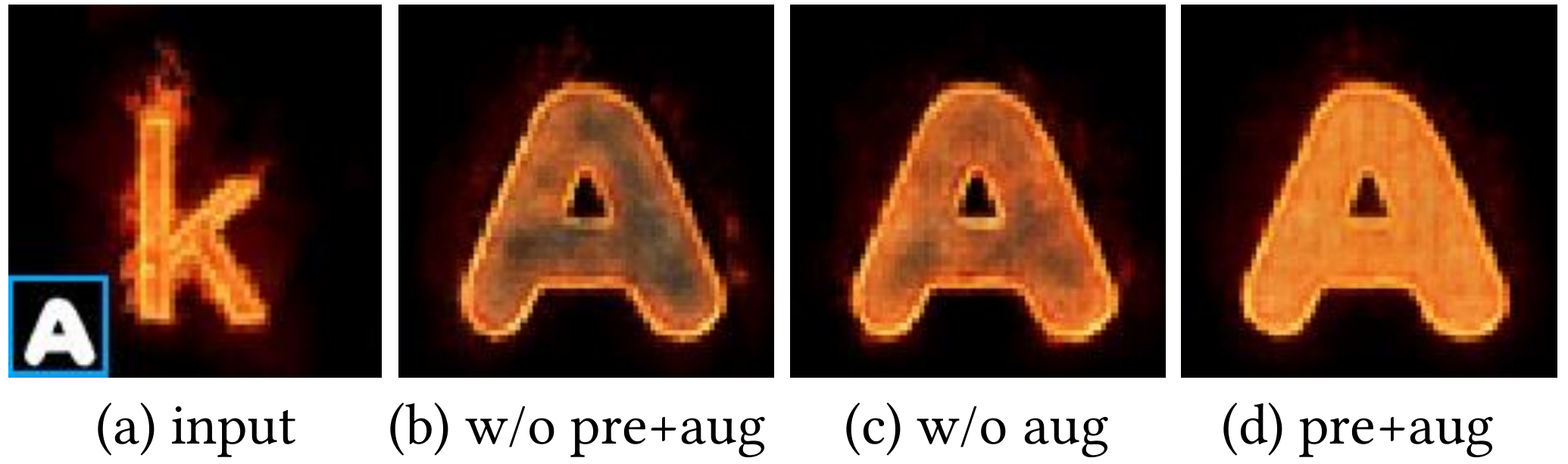}
  \caption{A comparison of results with and without our distribution-aware data preprocessing and augmentation.}\label{fig:ablation2}
\end{figure}

\begin{figure*}[t]
  \centering
  \subfigure[Visualization of VGG features of all text effects.]{
  \includegraphics[width=0.37\linewidth]{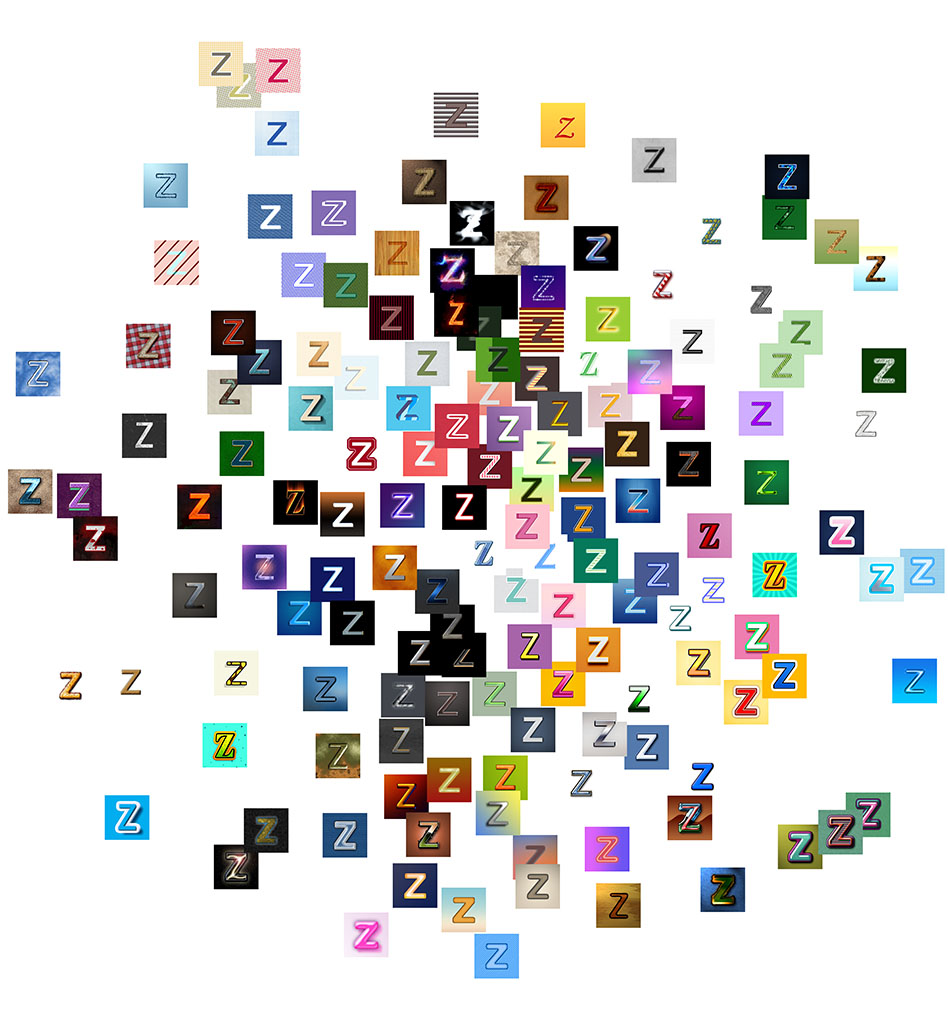}}
  \subfigure[Distribution of background, foreground, stroke, and stereo effects subclasses.]{
  \includegraphics[width=0.61\linewidth]{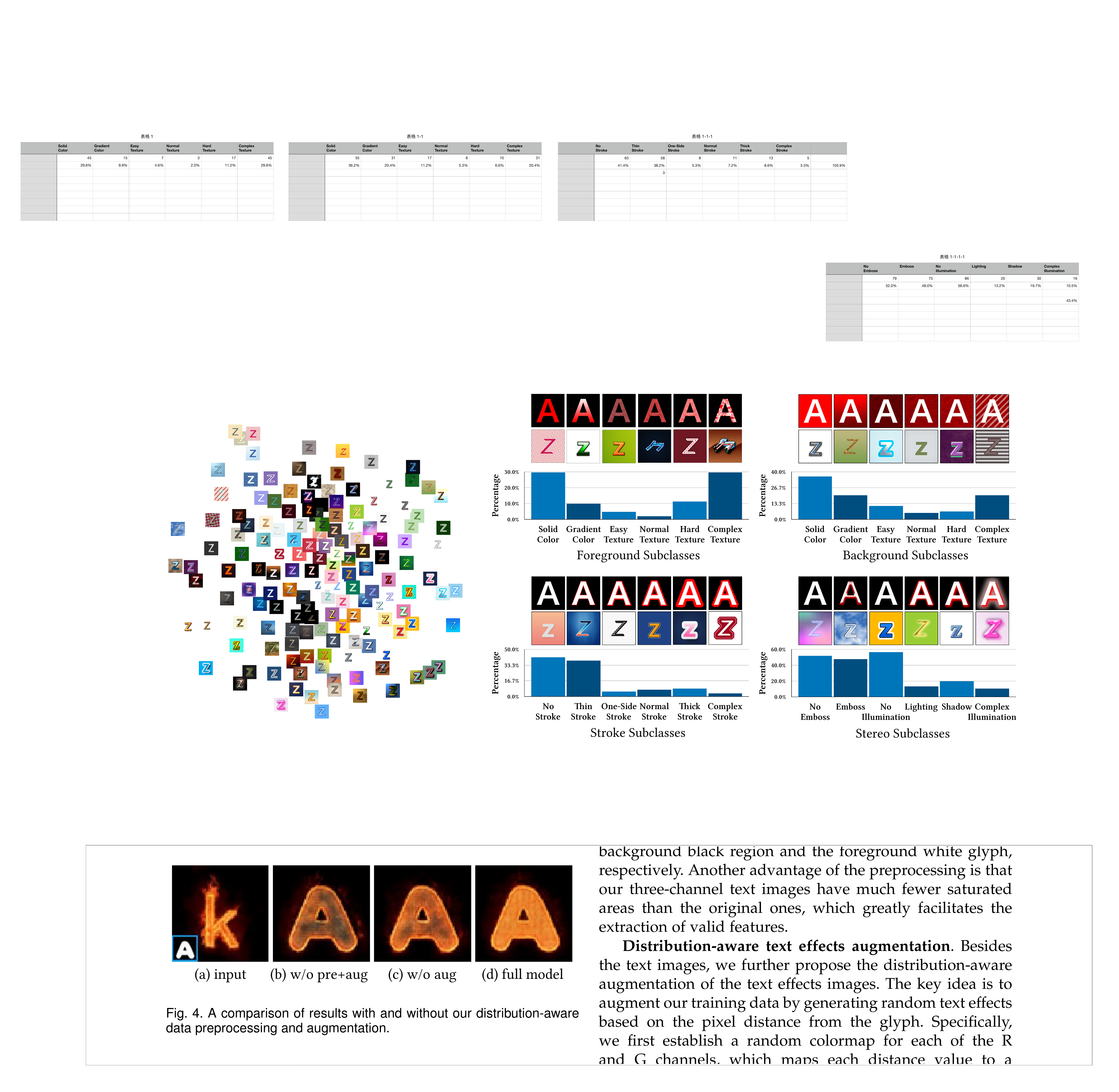}}
  \caption{Statistics of TE141K. (a) Visualized text effect distribution in TE141K by t-SNE~\cite{maaten2008visualizing} based on their VGG features~\cite{simonyan2014very}. Perceptually similar text effects are placed close to each other. The even distribution  indicates the diversity and richness of our dataset. (b) Distribution of different background, foreground, stroke, and stereo effects subclasses of our dataset. Representative images are shown at the top. The first row contains schematics where red is used to represent specific text effect subclasses. The second row contains representative samples from TE141K.}
  \label{fig:stat}
\end{figure*}

\textbf{Distribution-aware text image preprocessing}. As reported in~\cite{Yang2017Awesome}, the spatial distribution of the texture in text effects is highly related to its distance from the glyph, forming an effective prior for text effect transfer. To leverage this prior, we propose a distribution-aware text image preprocessing to directly feed models trained on TE141K with distance cues. As shown in Fig.~\ref{fig:preprocessing}, we extend the raw text image from one channel to three channels. The R channel is the original text image, while the G channel and B channel are distance maps where the value of each pixel is its distance to the background black region and the foreground white glyph, respectively. Another advantage of the preprocessing is that our three-channel text images have much fewer saturated areas than the original ones, which greatly facilitates the extraction of valid features.

\textbf{Distribution-aware text effect augmentation}. In addition to the text images, we further propose the distribution-aware augmentation of the text effect images. The key idea is to augment our training data by generating random text effects based on the pixel distance from the glyph. Specifically, we first establish a random colormap for each of the R and G channels, which maps each distance value to a corresponding color. Then, we use the colormaps of the R and G channels to tint the background black region and the foreground white glyph in the text image separately.
Figs.~\ref{fig:preprocessing}(c)-(e) show examples of the randomly generated text effect images. With colors that reflect structural distribution, these images can effectively guide transfer models to identify the spatial relationship between the text effects and the glyphs. In addition, data augmentation could also increase the generalization capabilities of the model.

In Fig.~\ref{fig:ablation2}, we examine the effect of our distribution-aware text image preprocessing and text effect augmentation on TE141K through a comparative experiment with the model proposed in Section~\ref{sec:tetgan}.
Without preprocessing and augmentation, the inner flame textures are not synthesized correctly.
As seen in Fig.~\ref{fig:ablation2}(d), our distribution-aware data augmentation strategy helps the network learn to infer textures based on their correlated position on the glyph and synthesize better flame textures.

\subsection{Dataset Statistics}
\label{sec:stat}

In the proposed dataset, there are a wide variety of text effects. To explore their distribution, we first obtain their feature maps from the fifth layer of the VGG network~\cite{simonyan2014very}. Then, we conduct nonlinear dimensionality reduction by t-SNE~\cite{maaten2008visualizing} and visualize the distribution of text effects. We choose VGG features instead of RGB information because text effects with similar structural characteristics are close to each other in VGG features. As illustrated in Fig.~\ref{fig:stat}(a), text effects are evenly distributed as a circle, indicating the diversity and richness of the text effects in our dataset.

Text effects usually consist of multiple fundamental visual elements, such as textures, stroke outlines and stereo effects. These factors may vary greatly and make text effect transfer a challenging task. To better investigate how they affect the model performance, we quantify these factors and manually label the text effects with the corresponding factors. We will introduce the statistics of these factors in this section and investigate their influence on the model performance in Section~\ref{sec:experiment_ana}.

We consider texture in two categories: foreground and background.
For background texture, based on complexity level, we classify it into 6 subclasses: \textit{Solid Color}, \textit{Gradient Color}, \textit{Easy Texture}, \textit{Normal Texture}, \textit{Hard Texture} and \textit{Complex Texture}.
\textit{Solid Color} is the simplest; transfer models only need to directly copy the original color.
\textit{Gradient Color} is more complex; the distribution of color is directional.
For the other four subclasses, there are textures on the background. In \textit{Easy Texture}, textures are imperceptible to human eyes. From \textit{Normal Texture} to \textit{Hard Texture} and \textit{Complex Texture}, textures become distinct and irregular. As shown in Fig.~\ref{fig:stat}(b), 36\% of the background is solid, and 20\% is of gradient color, which is consistent with the fact that background is often set to be simple to better emphasize the foreground glyph.
Similarly, we classify foreground into the same 6 subclasses. In contrast to the background, 30\% of the foreground has \textit{Complex Texture}. This may be because the foreground is the main body of text effects; therefore, it is often fully decorated to increase artistic quality.

\begin{table*}[t]
    \begin{center}
    \caption{Summary of benchmarking representative style transfer models and the proposed TET-GAN, showing the model type, model names, number of style supported per model ($\#$Style), support for supervised one-reference (SOF) or unsupervised one-reference (UOF) style transfer, availability of feed-forward fast style transfer, target type (general image style or text effect transfer), usage of deep models and the style representation.}
    \label{table:related_works}
    \begin{tabular*}{\textwidth}{clcccccl}
    \toprule
    \multirow{2}{*}{Type} & \multirow{2}{*}{Model} & \multicolumn{2}{c}{Flexibility} & Efficiency & \multicolumn{3}{c}{Model Design}\\
    \cmidrule(lr){3-4}\cmidrule(lr){5-5}\cmidrule(lr){6-8} & & $\#$Style & SOF/UOF & Feed-Forward & Type & Deep & Style Representation\\
    \midrule
    \multirow{3}{*}{Global} & NST~\cite{gatys2016image} & $\infty$ & UOF & $\times$ & general & $\surd$ & Gram matrix of deep features\\
     & AdaIN~\cite{huang2017adain} & $\infty$ & UOF & $\surd$ & general & $\surd$ & mean and variance of deep features\\
     & WCT~\cite{li2017universal} & $\infty$ & UOF & $\surd$ & general & $\surd$ & mean and covariance of deep features\\
    \midrule
    \multirow{6}{*}{Local} & Analogy~\cite{Hertzmann2001Image} & $\infty$ & SOS & $\times$ & general & $\times$ & image patches \\
    & Quilting~\cite{Efros2001Image} & $\infty$ & UOF & $\times$ & general & $\times$ & image patches\\
    & CNNMRF~\cite{Li2016Combining} & $\infty$ & UOF & $\times$ & general & $\surd$ & feature patches\\
    & Doodles~\cite{Champandard2016Semantic} & $\infty$ & SOF & $\times$ & general & $\surd$ & feature patches\\
    & T-Effect~\cite{Yang2017Awesome} & $\infty$ & SOF & $\times$ & text & $\times$ & image patches\\
    & UT-Effect~\cite{Yang2018Context} & $\infty$ & UOF & $\times$ & text & $\times$ & image patches\\
    \midrule
    \multirow{6}{*}{GAN-based} & Pix2pix~\cite{Isola2017Image} & 1/N* & $\times$ & $\surd$ & general & $\surd$ & learned features\\
    & BicycleGAN~\cite{zhu2017toward} & 1/N* & $\times$ & $\surd$ & general & $\surd$ & learned features\\
    & StarGAN~\cite{Choi2017StarGAN} & $N$ & $\times$ & $\surd$ & general & $\surd$ & learned features\\
    & MC-GAN~\cite{Azadi2017Multi} & $\infty$ & $\times$ & $\surd$ & text & $\surd$ & learned features\\
    \cmidrule(lr){2-8}
    & TET-GAN (ours) & $N$ & $\times$ & $\surd$ & text & $\surd$ & disentangled content and style features\\
    & TET-GAN+ (ours) & $\infty$ & SOF/UOF & $\surd$ & text & $\surd$ & disentangled content and style features\\
    \bottomrule
    \end{tabular*}
    \end{center}
    \raggedright \textit{* Pix2pix and BicycleGAN were originally designed to support only one style per model. In our experiment, we add an extra conditional text-style pair to their original input, which enables them to handle multiple styles in one model.\\ Note: Compared to TET-GAN, TET-GAN+ additionally makes use of the proposed one-reference finetuning strategy for style extension.}
\end{table*}

Stroke outlines and stereo effects are two common elements used in cartoon and 3D text effects, respectively. Based on complexity and thickness, we classify strokes into 6 subclasses: \textit{No Stroke}, \textit{Thin Stroke}, \textit{One-Side Stroke}, \textit{Normal Stroke}, \textit{Thick Stroke} and \textit{Complex Stroke}. We observe that the thickness of many \textit{Thin Stroke} and \textit{Normal Stroke} effects is uneven in direction; therefore, we further divide them into a more specific class, \textit{One-Side Stroke}. In TE141K, 59\% of the text effects contain strokes.
Stereo effects are usually a combination of multiple special effects, such as emboss and illumination effects. In addition, illumination effects can be further classified into lighting and shadow. Similarly, we classify stereo effects into 6 subclasses: \textit{No Emboss}, \textit{Emboss}, \textit{No Illumination}, \textit{Lighting}, \textit{Shadow}, and \textit{Complex Illumination}, where \textit{Complex Illumination} is a combination of more than two illumination effects. In TE141K, 48\% of the text effects contain embossing, and 43\% contain illumination effects.

\subsection{Comparison with Existing Datasets}

Datasets play an important role in the development of neural networks.
To the best of our knowledge, the dataset provided in the work on MC-GAN~\cite{Azadi2017Multi} is the only text effect dataset in the literature. To train MC-GAN, the authors collected 35 different kinds of text effects from the Internet.
For each style, only an extremely limited set of 26 images of capital letters with a small size of $64\times64$ are rendered, forming a total of 910 style images, which cannot support training a network that is robust enough to produce high-resolution images of arbitrary glyphs.
Therefore, MC-GAN can only handle 26 capital letters with a low resolution.
In contrast, our TE141K contains 152 different kinds of text effects. For each style, at least 837 glyphs are rendered, adding up to 141,081 image pairs in total. In addition, the image size reaches $320\times320$.
The proposed dataset exceeds that of~\cite{Azadi2017Multi} in terms of both quantity and diversity, supporting transfer models to render exquisite text effects on various glyphs.

\subsection{Text Effect Transfer Tasks}

On TE141K, we design three text effects transfer tasks according to the amount of information provided to transfer models, which will be benchmarked in Section~\ref{sec:experiment}:
\begin{itemize}
\setlength{\itemsep}{0pt}
\setlength{\parsep}{0pt}
\setlength{\parskip}{0pt}
\item \textbf{General Text Effect Transfer.} In this task, text styles in the training and testing phases are the same. Benchmarking is conducted with all three dataset subsets. During testing, the models are provided with an input example text effect image, its glyph counterpart and the target glyph. This task is relatively easy, since models can become familiar with text effects through a large amount of training data. The challenge lies in transferring multiple styles in one model and generalizing to unseen glyphs.
\item \textbf{Supervised One-Reference Text Effect Transfer.} In this task, text effects in the training and testing phases are different. For data-driven models, only the training sets of TE141K-E and TE141K-C are provided, and benchmarking is conducted on the testing set of TE141K-S. This task is more difficult, since models have to learn new text effects with only one example pair.
\item \textbf{Unsupervised One-Reference Text Effect Transfer.} This task is similar to supervised one-reference text effect transfer except that during testing, the glyph image of the example text effect image is not provided. This task is the most difficult, since transfer models have to distinguish the foreground and background by themselves.
\end{itemize}

\section{Benchmarking Existing Style Transfer Models}
\label{sec:related_work}

In this section, we briefly introduce existing representative style transfer models, which will be benchmarked on the proposed TE141K dataset in Section~\ref{sec:experiment}.
These models can be categorized based on their style representations, \textit{i.e.}, global statistics, local patches and learned features.
The choice of style representation can largely affect the characteristics of the model in terms of flexibility and efficiency.
To give an intuitive comparison, we summarize the models and their characteristics in Table~\ref{table:related_works}.

\textbf{Global models}.
Global models represent image styles as global statistics of image features and transfer styles by matching the global feature distributions between the style image and the generated image.
The advantage of explicitly defined style representation is that any input style can be modelled and transferred without requiring a large paired dataset, which is suitable for the task of unsupervised one-reference transfer.
\begin{itemize}[leftmargin=1.5em]
\item \textit{Neural Style Transfer (NST)}: The trend of parametric deep-based style transfer began with the pioneering work of neural style transfer~\cite{gatys2016image}. In NST, Gatys~\textit{et al.} formulated image styles as the covariance of deep features in the form of a Gram matrix~\cite{gatys2015texture} and transferred style by matching high-level representations of the content image and the Gram matrices; this technique demonstrates the remarkable representative power of convolutional neural networks (CNN) to model style. The main drawback of NST is its computationally expensive optimization procedure. Follow-up work~\cite{Johnson2016Perceptual,ulyanov2016texture,Wang2016Multimodal,Chen2017StyleBank} has been proposed to speed up NST by training a feed-forward network to minimize the loss of NST. However, efficiency is achieved at the expense of flexibility and quality. Thus, we select NST as our benchmarking global model.
\item \textit{Arbitrary Style Transfer (AdaIN)}: AdaIN~\cite{huang2017adain} presents a feature transformation framework, where the style is represented by the mean and variance of deep features. By aligning the statistics of the content features with those of the style features via adaptive instance normalization (AdaIN), AdaIN allows for fast arbitrary style transfer, achieving flexibility and efficiency simultaneously.
\item \textit{Universal Style Transfer (WCT)}: WCT~\cite{li2017universal} follows the feature transformation framework and represents style as the covariance of deep features, which can be adjusted by whitening/coloring transforms~(WCT). Compared to the variance in AdaIN~\cite{huang2017adain}, covariance can better capture high-level representations of the style.
\end{itemize}

\textbf{Local models}.
Local models represent styles as local patches and transfer styles by rearranging style patches to fit the structure of the content image.
Similar to global models, local models are capable of extracting style from only a single style image, and are therefore suitable for the task of one-reference transfer. Compared to global models, local patches better depict style details. However, matching patches can be time-consuming.
\begin{itemize}[leftmargin=1.5em]
\item \textit{Image Analogy (Analogy)}: Hertzmann~\textit{et al.} first presents a supervised framework called image analogy~\cite{Hertzmann2001Image}, which aims to learn the transformation between an unstylized source image and the corresponding stylized image. Style transfer is realized by applying the learned transformation to the target image. In \cite{Hertzmann2001Image}, the transformation is realized by replacing unstylized image patches with the corresponding stylized ones.
\item \textit{Image Quilting (Quilting)}: Image quilting~\cite{Efros2001Image} rearranges image patches from the style image according to the intensity or gradient of the content image, which can transfer style in an unsupervised manner.
\item \textit{CNNMRF}: CNNMRF~\cite{Li2016Combining} combines CNN with an MRF regularizer and models image style by local patches of deep features; it is suitable for fine structure preservation and semantic matching. However, it fails when the content and style images have strong semantic differences due to patch mismatches.
\item \textit{Neural Doodles (Doodle)}: Neural doodles~\cite{Champandard2016Semantic} introduce a supervised version of CNNMRF~\cite{Li2016Combining} by incorporating the semantic map of the style image as guidance, which alleviates the problem of patch mismatches.
\item \textit{Text Effect Transfer (T-Effect)}: Yang~\textit{et al.} proposed the first text effect transfer method, named T-Effect~\cite{Yang2017Awesome}. The authors modelled the text style by both the appearance of the patches and their correlated positions on the glyph, which achieved spatial consistency in text style.
\item \textit{Unsupervised Text Effect Transfer (UT-Effect)}: UT-Effect~\cite{Yang2018Context} is an unsupervised version of T-Effect~\cite{Yang2017Awesome}, which generates new text effects from texture images rather than given text effects. It further exploits the saliency of the textures to enhance the legibility of the rendered text.
\end{itemize}

\textbf{GAN-based models}.
GAN-based models are tasked to map between two or more domains; during this process, the style representation is implicitly learned from the data. This task-specific style representation has the advantage of producing vivid results but needs a large dataset for training and is usually hard to extend to new styles.
\begin{itemize}[leftmargin=1.5em]
\item \textit{Pix2pix-cGAN (Pix2pix)}: Pix2pix~\cite{Isola2017Image} presents a general-purpose framework built upon U-Net~\cite{Ronneberger2015U} and PatchGAN~\cite{Isola2017Image} for the task of image-to-image translation and has shown high performance in many applications, such as style transfer, colorization, semantic segmentation and daytime hallucination. Despite its high performance, Pix2pix~\cite{Isola2017Image} is designed for two domains and thus has limited flexibility in handling multiple styles.
\item \textit{BicycleGAN}: BicycleGAN~\cite{zhu2017toward} tackles the problem of generating diverse outputs by encouraging bijective consistency between the output and learned latent features. However, it is ineffective in multistyle transfer.
\item \textit{StarGAN}: StarGAN~\cite{Choi2017StarGAN} utilizes additional one-hot vectors as input to specify the target domain so that the network can learn a mapping between multiple domains, which is more flexible.
\item \textit{Multi-Content GAN (MC-GAN)}: Azadi~\textit{et al.}~\cite{Azadi2017Multi} presented an MC-GAN for the stylization of capital letters, which combines font transfer and text effect transfer using two successive subnetworks. A leave-one-out approach is introduced for few-reference style transfer (it takes about 6 example images as input, as reported in~\cite{Azadi2017Multi}), making the model more flexible. However, MC-GAN is designed to handle only $26$ capital letters with a small image resolution of $64\times64$, which highly limits its uses.
\end{itemize}

\begin{figure*}[htbp]
  \centering
  \includegraphics[width=1\linewidth]{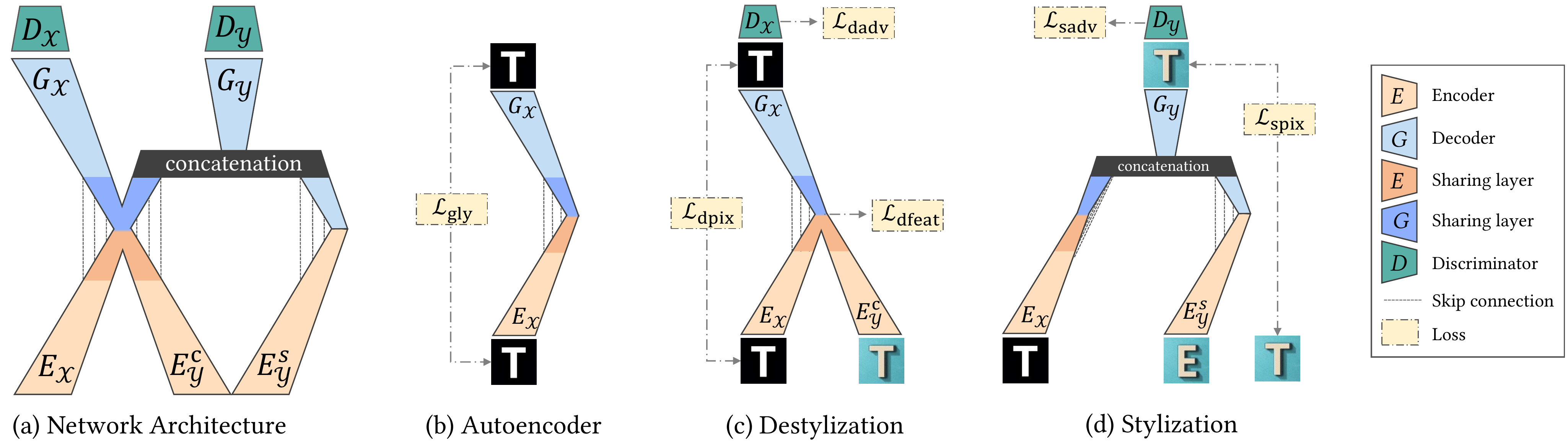}
  \caption{The TET-GAN architecture. (a) An overview of the TET-GAN architecture. Our network is trained via three objectives: an autoencoder, destylization and stylization. (b) A glyph autoencoder to learn content features. (c) Destylization by disentangling content features from text effect images. (d) Stylization by combining content and style features.}\label{fig:network-overview}
\end{figure*}

\section{TET-GAN for Text Effect Transfer}
\label{sec:tetgan}

To construct a baseline model for TE141K, we propose a deep-based approach named TET-GAN to disentangle and recombine the content and style features of text effect images so that it can simultaneously handle multiple styles in TE141K. We further propose a one-reference finetuning strategy that flexibly extends TET-GAN to new styles.

\subsection{Network Architecture and Loss Function}

Our goal is to learn a two-way mapping between two domains $\mathcal{X}$ and $\mathcal{Y}$, which represent a collection of text images and text effect images, respectively.
As shown in Fig.~\ref{fig:network-overview}, TET-GAN consists of two content encoders $\{E_{\mathcal{X}},E^c_{\mathcal{Y}}\}$,
a style encoder $\{E^s_{\mathcal{Y}}\}$,
two domain generators $\{G_{\mathcal{X}},G_{\mathcal{Y}}\}$ and
two domain discriminators $\{D_{\mathcal{X}},D_{\mathcal{Y}}\}$.
$E_{\mathcal{X}}$ and $E^c_{\mathcal{Y}}$ map text images and text effect images, respectively, onto a shared content feature space, while $E^s_{\mathcal{Y}}$ maps text effect images onto a style feature space.
$G_{\mathcal{X}}$ generates text images from the encoded content features.
$G_{\mathcal{Y}}$ generates text effect images conditioned on both the encoded content and style features.
Based on the assumption that domains $\mathcal{X}$ and $\mathcal{Y}$ share a common content feature space,
we share the weights between the last few layers of $E_{\mathcal{X}}$ and $E^{c}_{\mathcal{Y}}$ as well as the first few layers of $G_{\mathcal{X}}$ and $G_{\mathcal{Y}}$.
Discriminators are trained to distinguish the generated images from the real ones.

TET-GAN is trained on three tasks:
\textbf{Glyph Autoencoder} $G_{\mathcal{X}}\circ E_{\mathcal{X}}:\mathcal{X}\rightarrow\mathcal{X}$ for learning content feature encoding,
\textbf{Destylization} $G_{\mathcal{X}}\circ E^c_{\mathcal{Y}}:\mathcal{Y}\rightarrow\mathcal{X}$ for learning content feature disentanglement from text effect images,
and \textbf{Stylization } $G_{\mathcal{Y}}\circ(E_{\mathcal{X}},E^s_{\mathcal{Y}}):\mathcal{X}\times\mathcal{Y}\rightarrow\mathcal{Y}$ for learning style feature disentanglement and combination with content features.
Therefore, our objective is to solve the min-max problem:
\begin{equation}\label{eq:total_loss}
  \min_{E,G}\max_{D} \mathcal{L}_{\text{gly}}+\mathcal{L}_{\text{desty}}+\mathcal{L}_{\text{sty}},
\end{equation}
where $\mathcal{L}_{\text{gly}}$, $\mathcal{L}_{\text{desty}}$ and $\mathcal{L}_{\text{sty}}$ are losses related to the glyph autoencoder, destylization, and stylization, respectively.

\textbf{Glyph Autoencoder}. First, the encoded content feature is required to preserve the core information of the glyph.
Thus, we impose an autoencoder $L_1$ loss to force the content feature to completely reconstruct the input text image:
\begin{equation}\label{eq:ae_rec_loss}
  \mathcal{L}_{\text{gly}}=\lambda_{\text{gly}}\mathbb{E}_{x}[\|G_{\mathcal{X}}(E_{\mathcal{X}}(x))-x\|_{1}].
\end{equation}

\textbf{Destylization}. For destylization, we sample from the training set a text-style pair $(x,y)$.
We would like to map $x$ and $y$ onto a shared content feature space, where the feature can be used to reconstruct $x$, leading to the $L_1$ loss:
\begin{equation}\label{eq:desty_pix_loss}
  \mathcal{L}_{\text{dpix}}=\mathbb{E}_{x,y}[\|G_{\mathcal{X}}(E^{c}_{\mathcal{Y}}(y))-x\|_{1}].
\end{equation}
Furthermore, we would like $E^{c}_{\mathcal{Y}}$ to approach the ideal content feature extracted from $x$.
To enforce this constraint, we formulate the feature loss as:
\begin{equation}\label{eq:desty_feat_loss}
  \mathcal{L}_{\text{dfeat}}=\mathbb{E}_{x,y}[\|S_{\mathcal{X}}(E^{c}_{\mathcal{Y}}(y))-z\|_{1}],
\end{equation}
where $S_{\mathcal{X}}$ represents the sharing layers of $G_{\mathcal{X}}$ and $z=S_{\mathcal{X}}(E_{\mathcal{X}}(x))$.
Our feature loss guides the content encoder $E^c_{\mathcal{Y}}$ to remove the style elements from the text effect image, preserving only the core information of the glyph.

Finally, we impose conditional adversarial loss to improve the quality of the generated results. $D_{\mathcal{X}}$ learns to determine the authenticity of the input text image and whether it matches the given text effect image. At the same time, $G_{\mathcal{X}}$ and $E^{c}_{\mathcal{Y}}$ learn to confuse $D_{\mathcal{X}}$:
\begin{equation}\label{eq:desty_gan_loss}
\begin{aligned}
  \mathcal{L}_{\text{dadv}}=&~\mathbb{E}_{x,y}[\log D_{\mathcal{X}}(x,y)]\\
  -&~\mathbb{E}_{y}[\log(1-D_{\mathcal{X}}(G_{\mathcal{X}}(E^{c}_{\mathcal{Y}}(y)),y))].
\end{aligned}
\end{equation}
The total loss for destylization takes the following form:
\begin{equation}\label{eq:desty_total_loss}
  \mathcal{L}_{\text{desty}}=\lambda_{\text{dpix}}\mathcal{L}_{\text{dpix}}+
  \lambda_{\text{dfeat}}\mathcal{L}_{\text{dfeat}}+\lambda_{\text{dadv}}\mathcal{L}_{\text{dadv}},
\end{equation}

\textbf{Stylization}. For the task of stylization, we sample from the training set a text-style pair $(x,y)$ and a text effect image $y'$ that shares the same style with $y$ but has a different glyph.
We first extract the content feature from $x$ and the style feature from $y'$, which are then concatenated and fed into $G_{\mathcal{Y}}$ to generate a text effect image to approximate the ground-truth $y$ in an $L_1$ sense:
\begin{equation}\label{eq:sty_pix_loss}
  \mathcal{L}_{\text{spix}}=\mathbb{E}_{x,y,y'}[\|G_{\mathcal{Y}}(E_{\mathcal{X}}(x),E^{s}_{\mathcal{Y}}(y'))-y\|_{1}],
\end{equation}
and confuse $D_{\mathcal{Y}}$ with conditional adversarial loss:
\begin{equation}\label{eq:sty_adv_loss}
\begin{aligned}
  \mathcal{L}_{\text{sadv}}=&~\mathbb{E}_{x,y,y'}[\log D_{\mathcal{Y}}(x,y,y')]\\
  -&~\mathbb{E}_{x,y'}[\log(1-D_{\mathcal{Y}}(x,G_{\mathcal{Y}}(E_{\mathcal{X}}(x),E^{s}_{\mathcal{Y}}(y')),y'))].
\end{aligned}
\end{equation}
Our final loss for stylization is:
\begin{equation}\label{eq:sty_total_loss}
  \mathcal{L}_{\text{sty}}=\lambda_{\text{spix}}\mathcal{L}_{\text{spix}}+\lambda_{\text{sadv}}\mathcal{L}_{\text{sadv}}.
\end{equation}

\subsection{One-Reference Text Effect Transfer}
\label{sec:fewshot}

As introduced in Section~\ref{sec:related_work}, GAN-based methods are by nature heavily dependent on datasets and usually require thousands of training images, which greatly limits their applicability.
To build a baseline that supports personalized style transfer, as the global and local methods do,
we propose an one-reference finetuning strategy for style extension, where only one example style image is required.

\begin{figure}[t]
  \centering
    \includegraphics[width=1\linewidth]{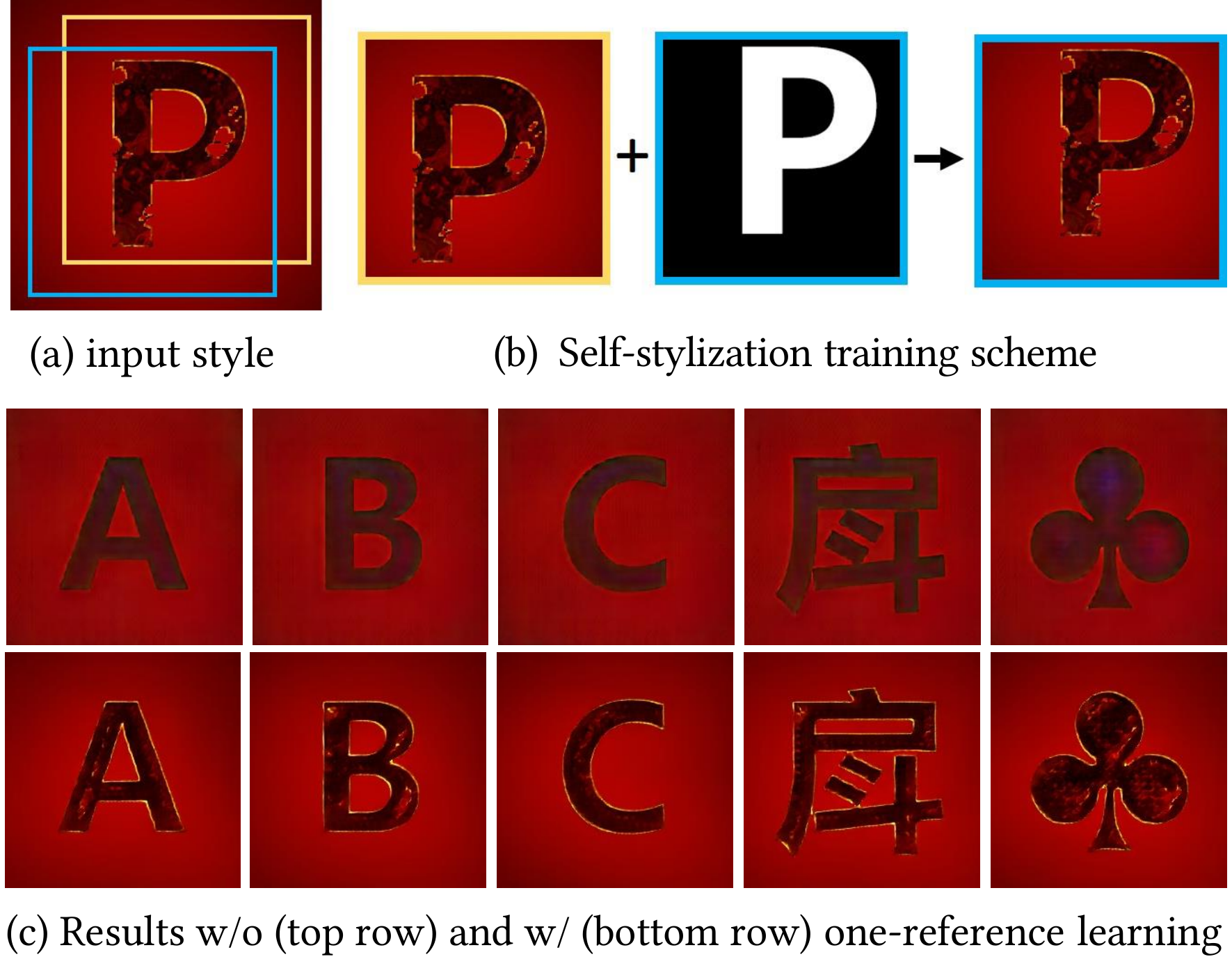}
  \caption{One-reference text effects transfer. (a) New, user-specified text effects.
  (b) Random crop of the style image to generate image pairs for training.
  (c) Top row: Stylization result on an unseen style. Bottom row: Stylization result after one-reference finetuning. The model finetuned over (a) is able to transfer text effects onto other unseen characters.}\label{fig:finetune}
\end{figure}

\textbf{One-reference supervised learning}. For an unseen style in Fig.~\ref{fig:finetune}(a), as shown in the top row of Fig.~\ref{fig:finetune}(c), TET-GAN trained on TE141K-C fails to synthesize texture details.
To solve this problem, inspired by self-supervised adversarial training~\cite{Yang2018nonstationary}, we propose a simple yet efficient ``self-stylization'' training scheme.
As shown in Fig.~\ref{fig:finetune}(b), we randomly crop the images to obtain many text effect patches that have the same style but differ in the pixel domain. In other words, $x$, $y$, and $y'$ in Eqs.~(\ref{eq:total_loss})-(\ref{eq:sty_total_loss}) are patches cropped from the given image pair.
They constitute a training set to finetune TET-GAN so that it can learn to reconstruct vivid textures, as shown in the bottom row of Fig.~\ref{fig:finetune}(c). Note that our model finetuned over a single image can generalize well to other very different glyphs.
Compared to the texture synthesis task~\cite{Yang2018nonstationary}, our one-reference style transfer task is more challenging, which requires generalization to glyphs.
Thus, our training scheme further requires the model to be pretrained on a large amount of supervised data to learn the domain knowledge of the glyphs. As we will show later in Fig.~\ref{fig:ablation3}, such domain knowledge learnt from TE141K plays an important role in improving the one-reference text effect transfer.

\textbf{One-reference unsupervised learning}. For an unseen style $y$ without a provided text image $x$, it is intuitive to exploit our destylization submodule to generate $x$ from $y$ and transform this one-reference unsupervised problem to a supervised one. In other words, $\tilde{x}=G_{\mathcal{X}}(E^{c}_{\mathcal{Y}}(y))$ is used as an auxiliary $x$ during finetuning.
Considering that the accuracy of the content features extracted from $\tilde{x}$ cannot be guaranteed, a style reconstruction loss is employed to further constrain the content features to help reconstruct $y$:
\begin{equation}\label{eq:unsuper_loss}
  \mathcal{L}_{\text{srec}}=\lambda_{\text{srec}}\mathbb{E}_{y}[\|G_{\mathcal{Y}}(E^{c}_{\mathcal{Y}}(y),E^{s}_{\mathcal{Y}}(y))-y\|_{1}].
\end{equation}
Our objective for unsupervised learning takes the form:
\begin{equation}\label{eq:unsuper_loss}
  \min_{E,G}\max_{D} \mathcal{L}_{\text{gly}}+\mathcal{L}_{\text{desty}}+\mathcal{L}_{\text{sty}}+\mathcal{L}_{\text{srec}}.
\end{equation}

The model of TET-GAN combining the one-reference finetuning strategy is named TET-GAN+.

\subsection{Semi-Supervised Text Effects Transfer}
\label{sec:semisupervised}

With the help of paired data provided by the proposed TE141K,
we can explore the potential of TET-GAN in semisupervised learning,
where the model is givensufficient but unpaired data for style extension. The main challenge is to establish an effective mapping between the style data and the glyph data.
Inspired by the adversarial augmentation proposed by~\cite{simo2018mastering},
we propose a hybrid supervised and unsupervised learning framework of TET-GAN,
where two augmentation discriminators $D^{\text{aug}}_{\mathcal{X}}$ and $D^{\text{aug}}_{\mathcal{Y}}$ are introduced to receive unlabeled data.
Specifically, $D^{\text{aug}}_{\mathcal{X}}$ is tasked to only judge the authenticity of the input text image without the need to determine whether it matches the given text effect image, and it uses the following objective function:
\begin{equation}\label{eq:desty_aug_loss}
\begin{aligned}
  \mathcal{L}_{\text{daug}}=&~\mathbb{E}_{x}[\log D^{\text{aug}}_{\mathcal{X}}(x)]\\
  -&~\mathbb{E}_{y}[\log(1-D^{\text{aug}}_{\mathcal{X}}(G_{\mathcal{X}}(E^{c}_{\mathcal{Y}}(y))))].
\end{aligned}
\end{equation}
The objective function of $D^{\text{aug}}_{\mathcal{Y}}$ is similarly defined:
\begin{equation}\label{eq:sty_aug_loss}
\begin{aligned}
  \mathcal{L}_{\text{saug}}=&~\mathbb{E}_{y,y'}[\log D^{\text{aug}}_{\mathcal{Y}}(y,y')]\\
  -&~\mathbb{E}_{x,y'}[\log(1-D^{\text{aug}}_{\mathcal{Y}}(G_{\mathcal{Y}}(E_{\mathcal{X}}(x),E^{s}_{\mathcal{Y}}(y')),y'))].
\end{aligned}
\end{equation}
The advantage is that the supervised data serve as an anchor to force the model to generate outputs consistent with the inputs, while the unsupervised data can teach the model to deal with a wider variety of glyph and style inputs.

\begin{figure}[t]
\centering
\includegraphics[width=1\linewidth]{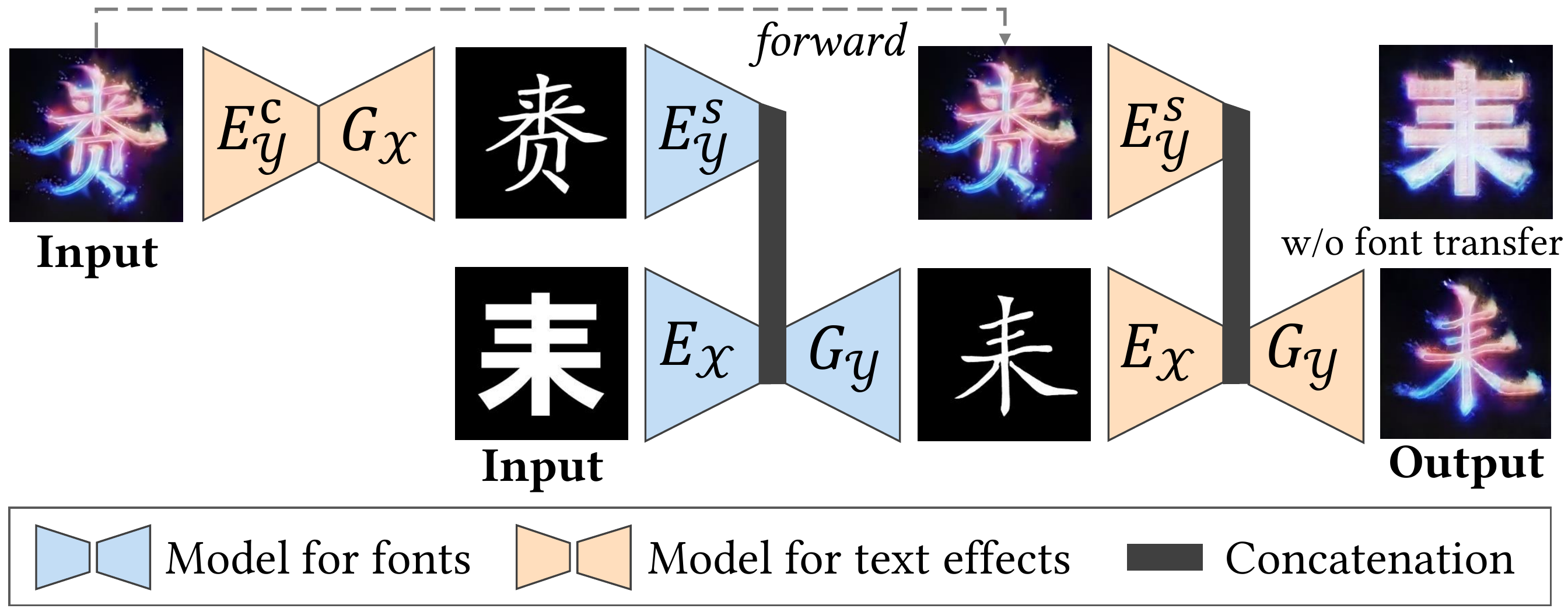}
\caption{Hybrid font style and text effect transfer framework. Two TET-GANs trained on the font dataset and text effects dataset constitute a uniform framework to transfer both font style and text effects.}
\label{fig:font-transfer-network}
\end{figure}

\subsection{Joint Font Style and Text Effect Transfer}

As we will show later, TET-GAN has a good generalizability across fonts and is capable of transferring the text effects on a reference image to other glyphs in different font styles.
However, some text effects are designed specifically for certain fonts. In Fig.~\ref{fig:font-transfer-network}, we show a case where the neon style suits a serif font (regular script) but not a sans-serif font (Microsoft Yahei).
Thus, it is a good option to match the font styles.
In this section, we explore the potential for TET-GAN in transferring the font style itself.
Specifically, we choose an anchor font $\mathcal{F}_0$ (in this paper, we use Microsoft Yahei) as an unstyled font.
All other fonts are regarded as stylized versions.
During training, $\{x,y,y'\}$ represents one character in the anchor font $\mathcal{F}_0$, the same character in another font $\mathcal{F}_1$, and another character in the other font $\mathcal{F}_1$.
All other training processes are the same as for learning text effects.

Then, two TET-GANs trained on font styles and text effects can constitute a uniform framework, as illustrated in Fig.~\ref{fig:font-transfer-network}. For an input style image, we first remove its text effects and obtain its raw text image, which is used as the reference font to adjust the input text. After that, we can use the original style image to render the text effects onto the font transfer result. The final output shares both the font style and text effects with the input style.

\begin{figure*}[t]
\centering
\includegraphics[width=1\linewidth]{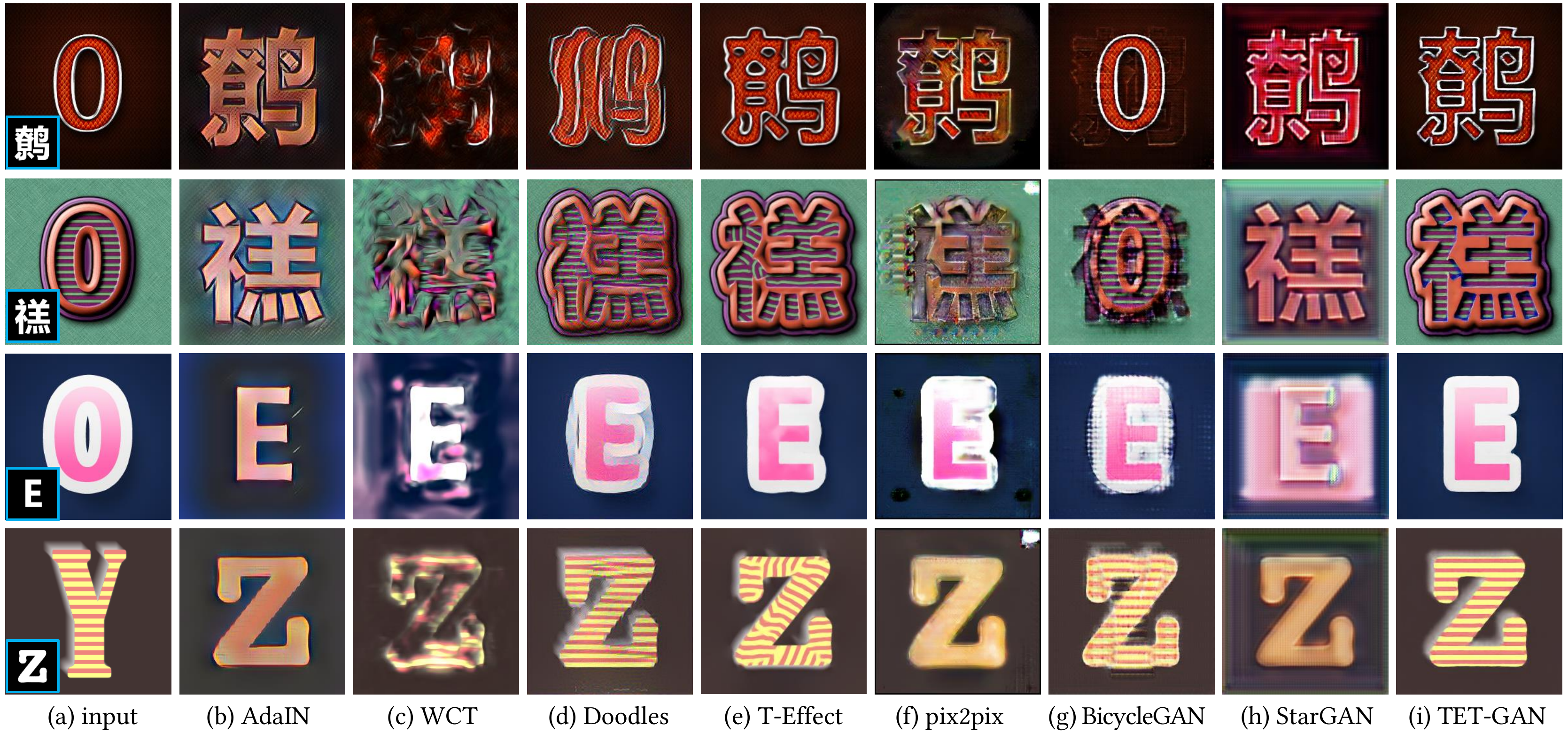}\vspace{-2mm}
\caption{
	Comparison with state-of-the-art methods on general text effect transfer.
	(a) Input example text effects with the target text in the lower-left corner.
	(b) AdaIN~\cite{huang2017adain}.
	(c) WCT~\cite{li2017universal}.
	(d) Doodles~\cite{Champandard2016Semantic}.
	(e) T-Effect~\cite{Yang2017Awesome}.
	(f) Pix2pix~\cite{Isola2017Image}.
	(g) BicycleGAN~\cite{zhu2017toward}.
	(h) StarGAN~\cite{Choi2017StarGAN}.
	(i) TET-GAN.
 }\vspace{-1mm}
\label{fig:comparison}
\end{figure*}

\section{Experimental Results}
\label{sec:experiment}

In this section, 15 state-of-the-art style transfer models, including the proposed TET-GAN and TET-GAN+, are tested on the proposed TE141K. The models are summarized in Table~\ref{table:related_works}.
Through the experimental results, we provide a comprehensive benchmark for text effect transfer, analyze the performance-influencing factors of TE141K, and demonstrate the superiority of the proposed models.

\subsection{Experimental Settings}

\textbf{Implementation Details}. For state-of-the-art models, we use their public codes and default parameters. Three data-driven methods, Pix2pix~\cite{Isola2017Image}, BicycleGAN~\cite{zhu2017toward}, and StarGAN~\cite{Choi2017StarGAN}, are trained on the proposed dataset with our data preprocessing and their own data augmentation. To allow Pix2pix and BicycleGAN to handle multiple styles, we change their inputs from a single text image to a concatenation of three images: the example text effect image, its glyph counterpart and the target glyph. The architecture details of TET-GAN are provided in the supplementary material.

\textbf{Evaluation Metrics}. Since there is currently no evaluation metric specially designed for text effects, we choose to use two traditional metrics, the peak signal-to-noise ratio (PSNR) and structural similarity index (SSIM), which are widely applied for image quality assessment, and two neural-network-based metrics, perceptual loss and style loss, which are commonly used in style transfer.

The PSNR is an approximation to human perception. Let $X$ be the image predicted by a model, and $Y$ be the ground-truth image. The PSNR is defined as:
\begin{equation}
\begin{split}
	\text{PSNR}(X, Y) = 20 \cdot log_{10} ( \frac{255} {||X-Y||^2_2}).
\end{split}
\end{equation}

Compared with the PSNR, SSIM is more sensitive to changes in structural information. Given $\mu_x$ and $\mu_y$, the average of $X$ and $Y$, respectively; $\sigma_x^2$ and $\sigma_y^2$, the variance of $X$ and $Y$, respectively; $\sigma_{xy}$, the covariance of $X$ and $Y$; and $c_1=6.5025$ and $c_2=58.5225$, two variables to stabilize division with a weak denominator, SSIM is defined as:
\begin{equation}
\begin{split}
	\text{SSIM}(X, Y) = \frac { (2 \mu_x \mu_y + c_1) (2 \sigma_{xy}+c_2) } { (\mu_x^2 + \mu_y^2 + c_1) (\sigma_x^2 + \sigma_y^2 + c_2) }.
\end{split}
\end{equation}

In neural style transfer~\cite{gatys2016image}, perceptual loss and style loss measure the semantic similarity and style similarity of two images, respectively. Let $\mathcal{F}_l$ be the feature map of the $l$-th layer of a pretrained VGG-19 network, with Gram matrix $\mathcal{G}(\mathcal{F}) = \mathcal{F}\mathcal{F}^\top$; then, perceptual loss and style loss are:
\begin{equation}
\begin{aligned}
	\text{Perceptual}(X, Y) &= \sum_{l} || \mathcal{F}_l(X) - \mathcal{F}_l(Y) ||_2^2, \\
	\text{Style}(X, Y) &= \sum_{l} || \mathcal{G}(\mathcal{F}_l(X)) - \mathcal{G}(\mathcal{F}_l(Y)) ||_2^2.
\end{aligned}
\end{equation}
To measure the results from both local and global perspectives, we choose five layers \texttt{relu1\_1}, \texttt{relu2\_1}, \texttt{relu3\_1}, \texttt{relu4\_1} and \texttt{relu5\_1}.

Furthermore, a user study was conducted in which 20 observers were asked to score the comprehensive transfer performance with 1 to 5 points (where 5 is the best). For each observer, we randomly selected 5 input pairs and asked them to score the corresponding results of all models. Finally, we collected 2,000 scores and calculated the average score for each model on each task.


\subsection{Benchmarking on General Text Effect Transfer}

\label{sec:experiment_general_transfer}

\begin{table}[t]
    \begin{center}
    \caption{Performance benchmarking on the task of general text effect transfer with PSNR, SSIM, perceptual loss, style loss, and the average score of the user study. The best score in each column is marked in bold, and the second best score is underlined.}
    \label{table:result_1}
    \begin{tabular}{llllll}
    \toprule
    Model & PSNR & SSIM & Perceptual & Style & User\\
    \midrule
    AdaIN~\cite{huang2017adain}	& 13.939 & 0.612 & 1.6147 & 0.0038  & 1.89\\
	WCT~\cite{li2017universal}  & 14.802  & 0.619	& 1.8626 & 0.0036 & 1.63\\
    Doodles~\cite{Champandard2016Semantic}	& 18.172 & 0.666	 & 1.5763 & 0.0031  & 2.98\\
	T-Effect~\cite{Yang2017Awesome} & \underline{21.402} & 0.793 & \underline{1.0918} & \underline{0.0020}  & \underline{4.19}\\
    Pix2pix~\cite{Isola2017Image} & 20.518	& 0.798 & 1.3940	 & 0.0032  & 3.08\\
    BicycleGAN~\cite{zhu2017toward} & 20.950 & \underline{0.803} & 1.5080	 & 0.0033  & 2.63\\
	StarGAN~\cite{Choi2017StarGAN} & 14.977 & 0.612 & 1.9144	 & 0.0045  & 2.09\\
    TET-GAN (ours)	& \textbf{27.626}	& \textbf{0.900} & \textbf{0.8250}	& \textbf{0.0014} & \textbf{4.62}\\
	\bottomrule
    \end{tabular}
    \end{center}
\end{table}

We first benchmark models on general text effect transfer, where text styles in the training and testing phases are the same. As shown in Table~\ref{table:result_1}, TET-GAN performs the best on all metrics. Representative results are shown in Fig.~\ref{fig:comparison}.

Two representative global models, AdaIN~\cite{huang2017adain} and WCT~\cite{li2017universal}, fail to perform well in both quantity and quality. AdaIN fails to reconstruct details and has color deviations, while WCT creates wavy artifacts and mingles the foreground and background.
This may be because AdaIN and WCT are designed to model the style as global statistics, and this representation is not suitable for text effect transfer.

We select two representative local models, Doodles~\cite{Champandard2016Semantic} and T-Effect~\cite{Yang2017Awesome}. Doodles fails to preserve global structure and suffers from artifacts. In Table~\ref{table:result_1}, T-Effect performs second best on four metrics. However, since the T-Effect processes patches in the pixel domain, it causes obvious color and structure discontinuity. In addition, when the edge of the input glyph differs greatly from the target glyph, T-Effect fails to adapt well to the new glyph.

Regarding GAN-based methods, both Pix2pix~\cite{Isola2017Image} and BicycleGAN~\cite{zhu2017toward} achieve a PSNR of over 20 and an SSIM of over 0.79.
However, they cannot eliminate the original input, leaving some ghosting artifacts.
StarGAN~\cite{Choi2017StarGAN} learns some color mappings but fails to synthesize texture details and suffers from distinct checkerboard artifacts, which also leads to low quantitative performance.
In summary, GAN-based methods can largely realize text effect transfer; however, the visual effect is not satisfactory, which may be due to the instability of GAN and limited network capability.

By comparison, our network learns valid glyph features and style features, thus precisely transferring text effects while preserving the glyph well.

\begin{figure}[t]
  \centering
  \includegraphics[width=0.98\linewidth]{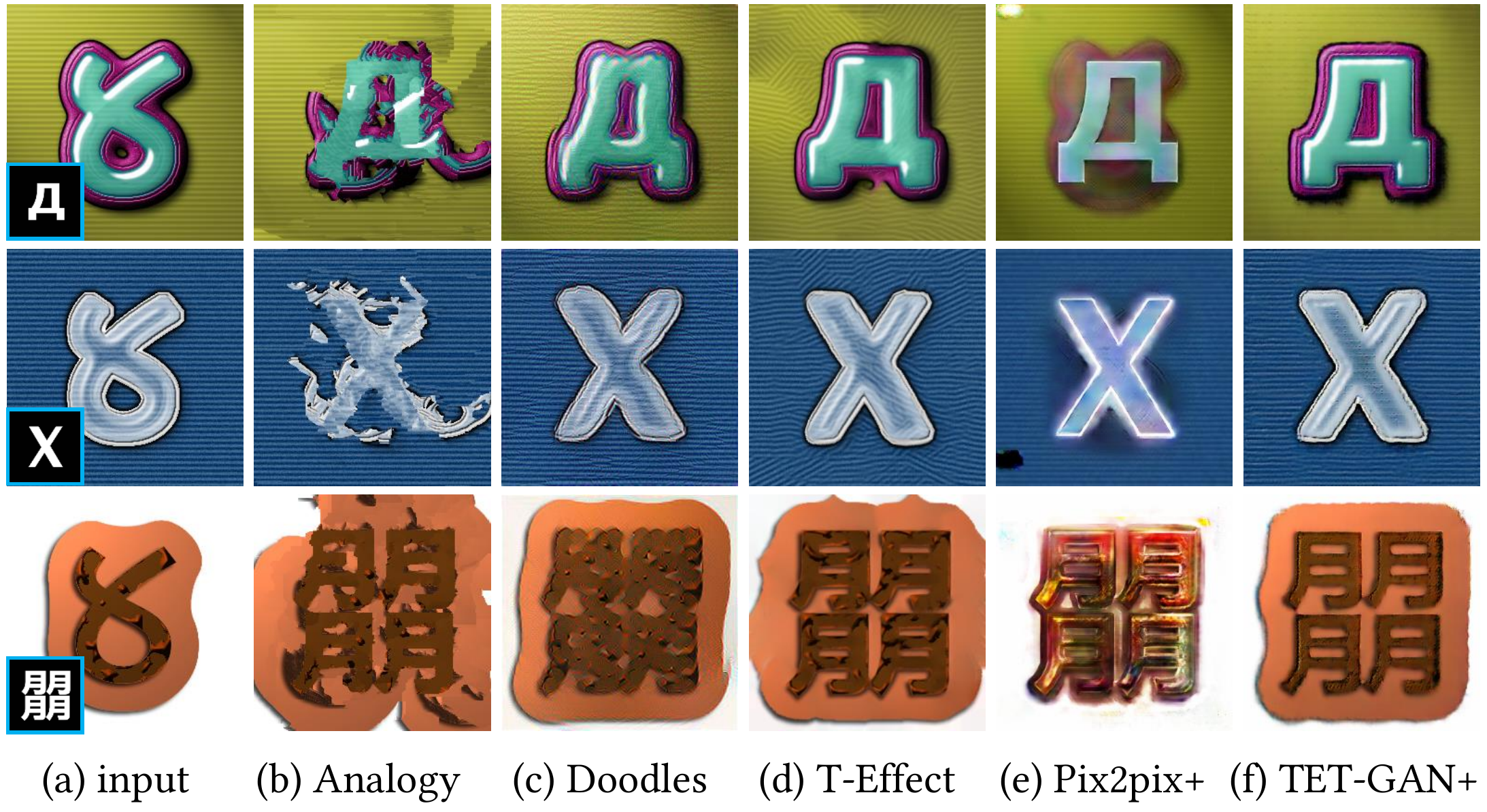}
\caption{
Comparison with other methods on one-reference supervised text effect transfer.
(a) Input example text effects with the target text in the lower-left corner.
(b) Analogy~\cite{Hertzmann2001Image}.
(c) Doodles~\cite{Champandard2016Semantic}.
(d) T-Effect~\cite{Yang2017Awesome}.
(e) Pix2pix+~\cite{Isola2017Image}.
(f) TET-GAN+.
}
 \label{fig:comparison2}
\end{figure}

\begin{figure*}[t]
  \centering
  \includegraphics[width=0.98\linewidth]{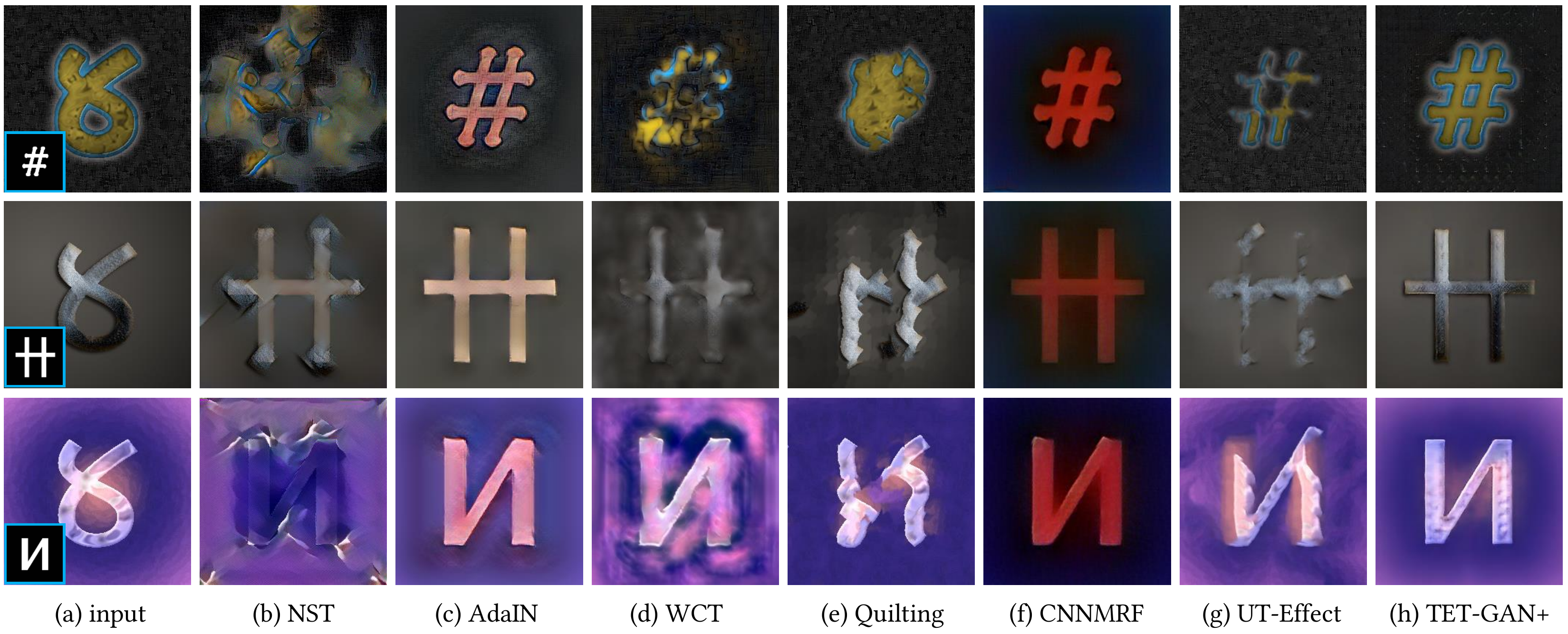}
\caption{
Comparison with other methods on one-reference unsupervised text effect transfer.
(a) Input example text effects with the target text in the lower-left corner.
(b) NST~\cite{gatys2016image}.
(c) AdaIN~\cite{huang2017adain}.
(d) WCT~\cite{li2017universal}.
(e) Quilting~\cite{Efros2001Image}.
(f) CNNMRF~\cite{Li2016Combining}.
(g) UT-Effect~\cite{Yang2018Context}.
(h) TET-GAN+.
}
\label{fig:comparison3}
\end{figure*}


\subsection{Benchmarking on Supervised One-Reference Text Effect Transfer}

In the task of supervised one-reference text effect transfer, only one observed example pair is provided.
In addition to existing methods, we select Pix2pix~\cite{Isola2017Image}, which has the best user score among the GAN-based models in the last task, and apply our one-reference learning strategy proposed in Section~\ref{sec:fewshot} to it, making Pix2pix able to process unseen text effects. This variant is named Pix2pix+.

As shown in Table~\ref{table:result_2}, the proposed TET-GAN+ achieves the best results on PSNR, SSIM, content loss and the user study, and performs second best on style loss.
T-Effect~\cite{Yang2017Awesome} performs slightly better than TET-GAN+ on style loss. This may be because style loss mainly focuses on local similarity, and the T-effect can easily achieve good local similarity by directly copying and fusing patches from the source image.
It can be observed in Fig.~\ref{fig:comparison2} that the T-Effect is less able to preserve shape, texture regularity and color continuity.
In terms of efficiency, the released T-Effect implemented in MATLAB requires approximately $150$ s per image with an Intel Xeon E5-1620 CPU (no GPU version available). In comparison, our feed-forward method only takes approximately $0.33$ s per image with an Intel Core i7-6850K CPU and $10$ ms per image with a GeForce GTX 1080 GPU after a three-minute finetuning.

\begin{table}[t]
    \begin{center}
    \caption{Performance benchmarking on the task of supervised one-reference text effect transfer with PSNR, SSIM, perceptual loss, style loss, and the average score of the user study. The best score in each column is marked in bold, and the second best score is underlined.}
    \label{table:result_2}
    \begin{tabular}{llllll}
    \toprule
    Model & PSNR & SSIM & Perceptual & Style & User\\
    \midrule
    Analogy~\cite{Hertzmann2001Image}  & 14.639 & 0.581  & 2.1202  & 0.0034 & 1.59\\
    Doodles~\cite{Champandard2016Semantic} & 17.653 & 0.636  & 1.6907  & 0.0028 & 3.20\\
    T-Effect~\cite{Yang2017Awesome}    & \underline{18.654}   & \underline{0.712}    & \underline{1.4023}    & \textbf{0.0022} & \underline{3.96}\\
    Pix2pix+~\cite{Isola2017Image}    & 16.656 & 0.660  & 1.7226  & 0.0037 & 2.30\\
    TET-GAN+ (ours)   & \textbf{20.192}    & \textbf{0.767} & \textbf{1.4017} & \underline{0.0026} & \textbf{4.26}\\
    \bottomrule
    \end{tabular}
    \end{center}
\end{table}

The other three methods are not as good as TET-GAN+ in terms of either quantitative or qualitative performance.
Two local methods face problems: Analogy~\cite{Hertzmann2001Image} cannot well preserve the shape of the glyph, while Doodles~\cite{Champandard2016Semantic} causes artifacts and distorts textures.
Although, taking advantage of the proposed finetuning strategy, Pix2pix+~\cite{Isola2017Image} can transfer basal colors to the target glyph, it suffers from severe detail loss. Compared with Pix2pix+, TET-GAN+ can adapt to and reconstruct new textures due to feature disentanglement.


\subsection{Benchmarking on Unsupervised One-Reference Text Effect Transfer}

\begin{table}[t]
    \begin{center}
    \caption{Performance benchmarking on the task of unsupervised one-reference text effect transfer with PSNR, SSIM, perceptual loss, style loss, and the average score of the user study. The best score in each column is marked in bold, and the second best score is underlined.}
    \label{table:result_3}
    \begin{tabular}{llllll}
    \toprule
    Model & PSNR & SSIM & Perceptual & Style & User\\
    \midrule
    NST~\cite{gatys2016image}  & 11.413 & 0.255  & 2.5705  & 0.0045 & 1.56\\
    AdaIN~\cite{huang2017adain}   & 13.443 & 0.579  & \underline{1.7342}    & 0.0033 & 1.92\\
    WCT~\cite{li2017universal} & 13.911 & 0.542    & 2.0934  & 0.0033 & 1.93\\
    Quilting~\cite{Efros2001Image} & 11.045 & 0.354  & 2.6733  & 0.0058 & 1.41\\
    CNNMRF~\cite{Li2016Combining}  & 09.309  & 0.337  & 1.8427  & 0.0042 & 1.79\\
    UT-Effect~\cite{Yang2018Context} & \underline{14.877} & \underline{0.609}  & 1.7551  & \underline{0.0028} & \underline{2.12}\\
    TET-GAN+ (ours)   & \textbf{18.724}    & \textbf{0.721} & \textbf{1.4933} & \textbf{0.0027} & \textbf{3.90}\\
    \bottomrule
    \end{tabular}
    \end{center}
\end{table}

The task of unsupervised one-reference text effect transfer is challenging; only one observed example is provided for the models. The advantages of our approach are more pronounced. As shown in Table~\ref{table:result_3}, TET-GAN+ achieves the best result on all five metrics. Representative results are illustrated in Fig.~\ref{fig:comparison3}.

Three global methods, NST~\cite{gatys2016image}, AdaIN~\cite{huang2017adain} and WCT~\cite{li2017universal}, all fail to produce satisfactory quantity and quality results. NST cannot correctly find the correspondence between the texture and glyph, leading to severe shape distortion and interwoven textures. AdaIN and WCT encounter the same problem as in Section~\ref{sec:experiment_general_transfer}. Concerning local methods, CNNMRF~\cite{Li2016Combining} fails to adapt text effect features to the target glyph. Quilting~\cite{Efros2001Image} and UT-Effect~\cite{Yang2018Context} distort glyphs. Taking full advantage of destylization, TET-GAN+ can distinguish the foreground and background well, and therefore can successfully reconstruct the global structure.

Compared with Table~\ref{table:result_2}, we find that the quantitative performance in Table~\ref{table:result_3} is much worse, which indicates that the unsupervised task is more difficult than the supervised one. This is obvious since the models are provided with much less information. However, the proposed TET-GAN+ still achieves satisfactory results, which demonstrates that the unsupervised task is solvable, and the proposed unsupervised one-reference training scheme is feasible.

\begin{figure}[t]
  \centering
  \includegraphics[width=\linewidth]{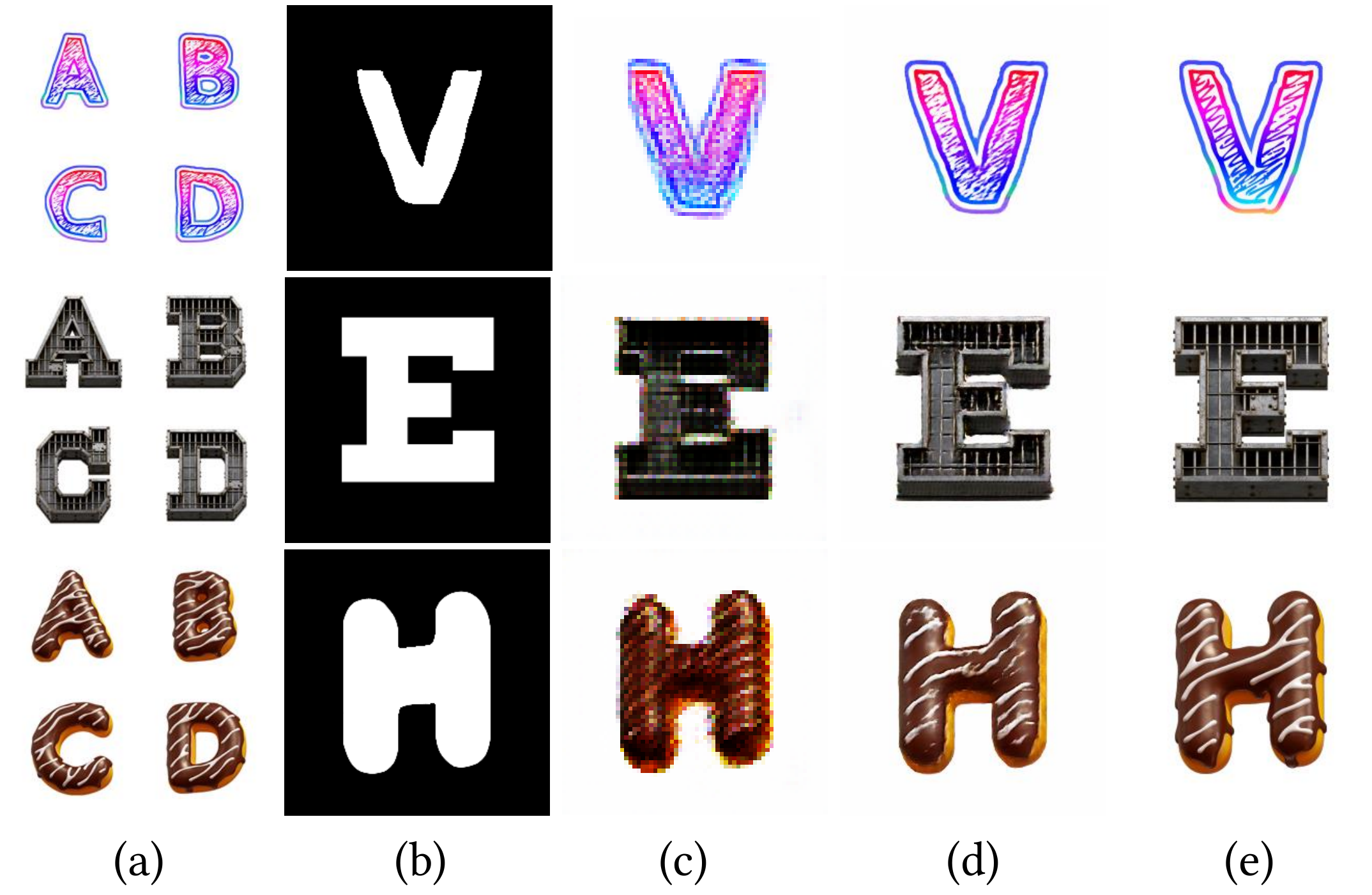}
  \caption{Comparison with MC-GAN~\cite{Azadi2017Multi} on TE141K-F. (a) Input style images for training. (b) Target glyph images. (c) Results of MC-GAN. (d) Results of TET-GAN+. (e) Ground-truth.}
\label{fig:mcgan}
\end{figure}

\subsection{Comparision with MC-GAN}

MC-GAN~\cite{Azadi2017Multi} is a few-reference method and can only handle 26 capital letters with a small image resolution of 64 $\times$ 64. Moreover, MC-GAN only supports text effects with white backgrounds. Therefore, to compare MC-GAN, we need to build a new testing set.

We first collect 10 styles from FlamingText\footnote{http://www6.flamingtext.com/All-Logos}, the same website used by MC-GAN. Note that FlamingText renders text effects in an automatic way, which is similar to our data collection method using PhotoShop batch tools. To fairly examine the ability of a model to handle unexpected text effects,
we collect 10 more challenging data from Handmadefont\footnote{https://handmadefont.com/}, an online shop of professionally designed handmade or 3D text effects. Finally, we collect 20 styles with 5 images for each style, summing up to 100 text effect images, and manually label the corresponding glyph for each style image. This testing set is named \textbf{TE141K-F} (`F' for few-reference).

Fig.~\ref{fig:mcgan} shows the comparison results where four characters are provided for few-reference learning and one for testing.
We directly feed MC-GAN with the ground-truth glyph images to make a fair comparison with TET-GAN.
As can be seen, the text effects in TE141K-F have quite challenging structures and textures, which are not reconstructed by MC-GAN.
In Fig.~\ref{fig:mcgan}(d), TET-GAN+ generates high-quality results with rich textures, and synthesizes complex but plausible structures. However, compared to the ground-truth, the details are still slightly distorted and mixed in the results of TET-GAN+.
This indicates substantial room for improvement in terms of few-reference learning.
Full results on TE141K-F are provided in the supplementary material.

\subsection{Analysis of TE141K}
\label{sec:experiment_ana}

We further analyze the performance-influencing factors of TE141K with the benchmarking results.
To quantify the difficulty of transferring a certain text style, we use its average user score value over eight transfer models benchmarked on the task of general text effect transfer to represent its transferring difficulty.

\begin{figure}[t]
  \centering
  \includegraphics[width=0.98\linewidth]{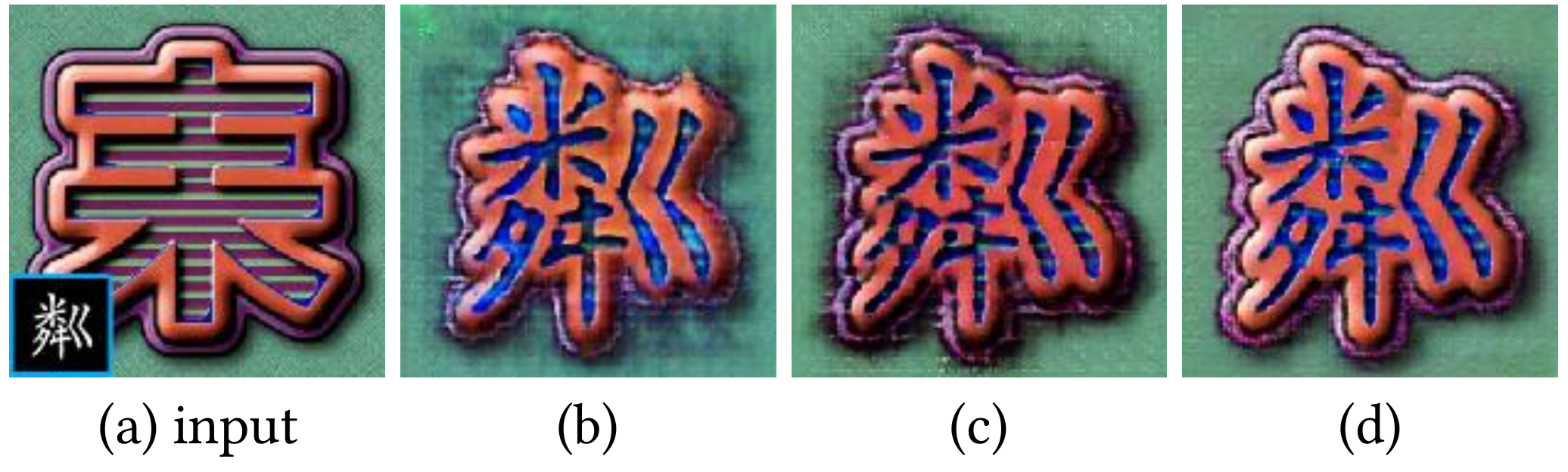}
  \caption{Effect of autoencoder loss and feature loss. (a) Input. (b) Model without $\mathcal{L}_{\text{gly}}$ and $\mathcal{L}_{\text{dfeat}}$. (c) Model without $\mathcal{L}_{\text{dfeat}}$. (d) Full model.}\label{fig:ablation1}
\end{figure}

\begin{figure}[t]
  \centering
  \includegraphics[width=0.98\linewidth]{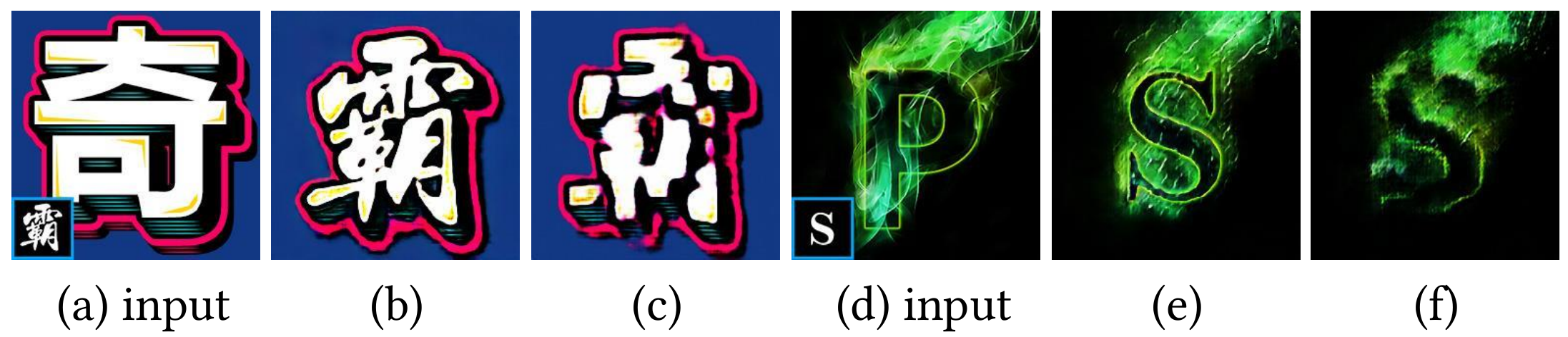}
  \caption{Effect of pretraining for one-reference text effect transfer. (a),(d): Input. (b),(e): Our results. (c),(f): Results without pretraining.}\label{fig:ablation3}
\end{figure}

To explore the factors that determine the difficulty of transferring text effects, we first use one-hot encoding to represent whether text effects contain the 24 subclasses introduced in Section~\ref{sec:stat}. Then, we adopt linear regression to fit the average user score. The Pearson correlation coefficient of this regression is $0.742$. The linear regression weights can indicate whether a subclass plays an important role in determining the user scores.
The three largest positive weight values are \textit{One-Side Stroke} (0.47798), \textit{Background Hard Texture} (0.39513), and \textit{Foreground Normal Texture} (0.37166). The reason for this is twofold. First, these distinct visual elements are easily perceived by human eyes to influence user decisions. Second, these visual elements are mostly anisotropic and irregular, which are difficult for transfer models to characterize and generate. Additionally, \textit{Normal Stroke} (-1.34573) has the lowest negative weight values, while the values of \textit{Thick Stroke} (-0.73353) and \textit{Thin Stroke} (-0.60621) are also negative. This is because with the help of glyph shape and data preprocessing to provide distance information, strokes become easy to model and reconstruct.
In conclusion, generating irregular textures or shapes around glyphs constitutes the major challenge of text effect transfer. This suggests focusing on modeling irregular textures or shapes in subsequent studies.

\begin{figure}[t]
  \centering
  \subfigure[Classification accuracy of content during training]{
  \includegraphics[width=0.9\linewidth]{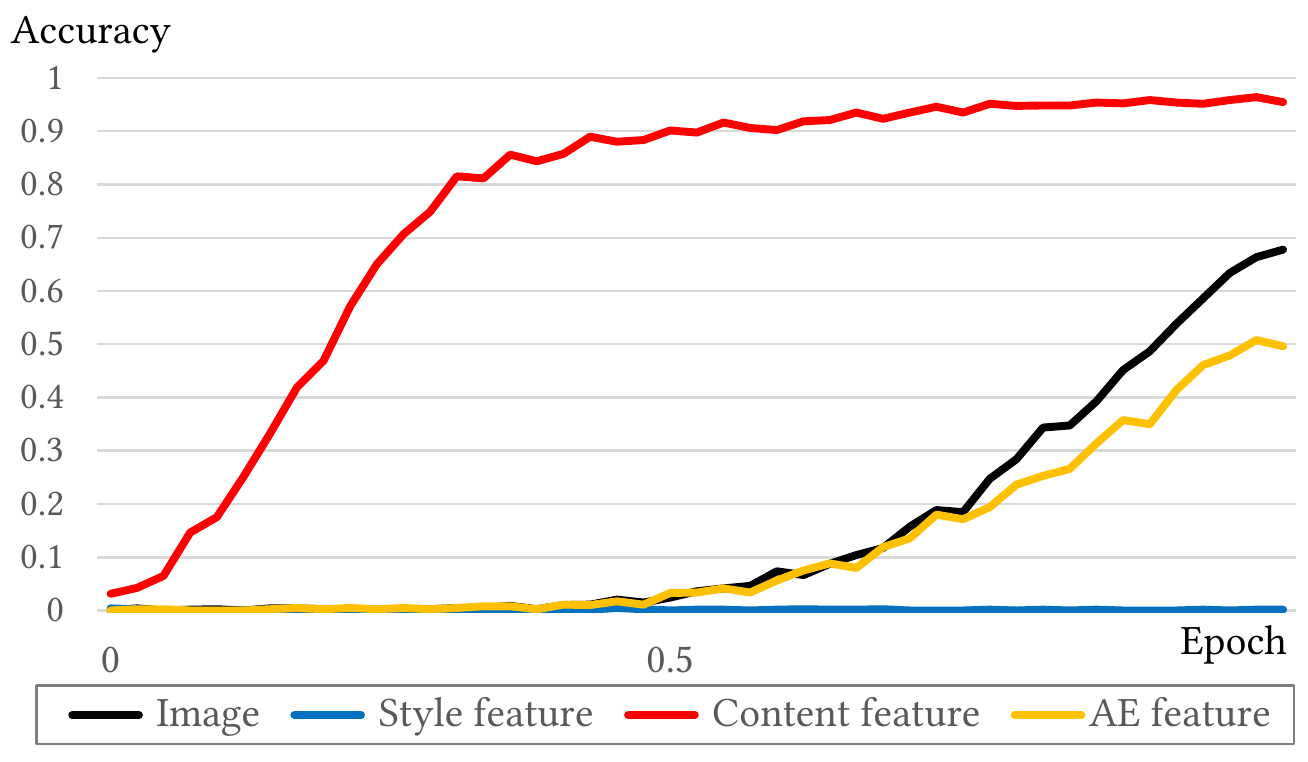}}
  \subfigure[Classification accuracy of styles during training]{
  \includegraphics[width=0.9\linewidth]{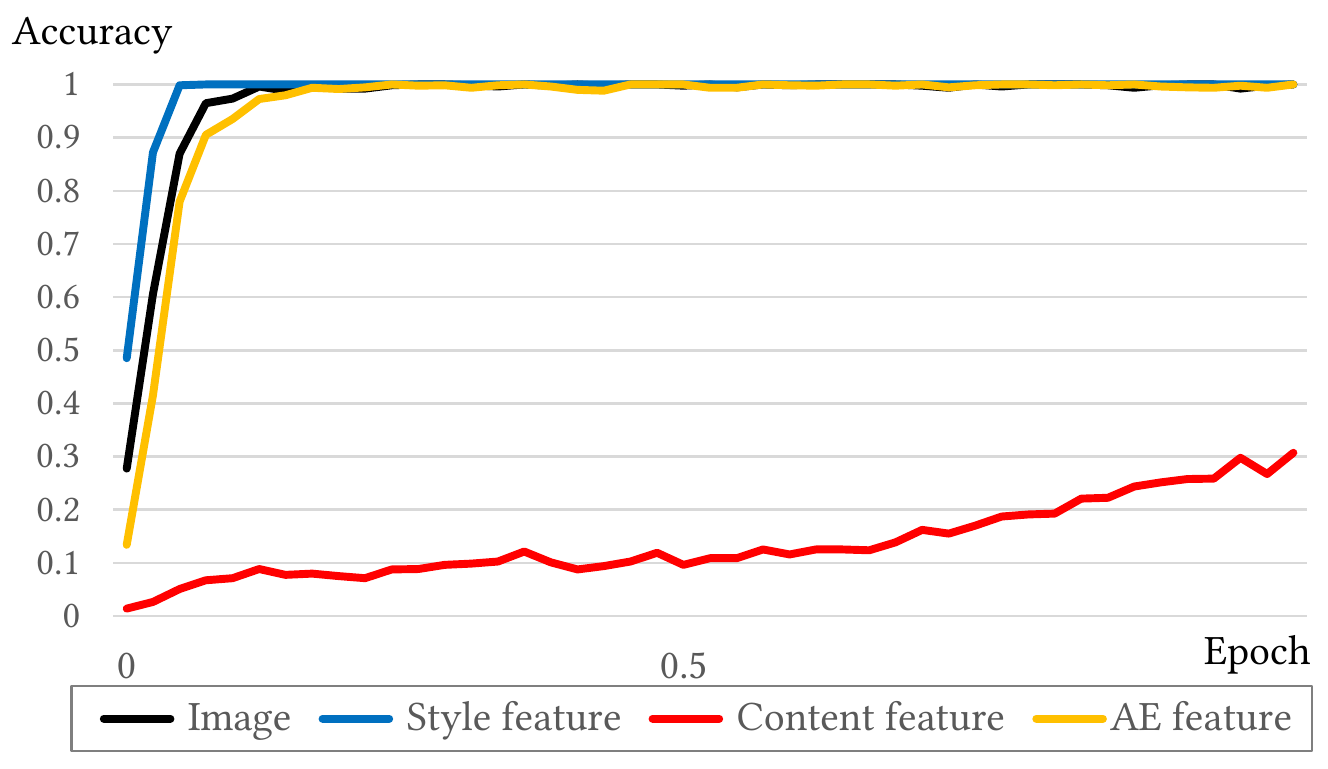}}
  \caption{Analysis of the disentangled style feature and content feature in text effect classification and glyph classification.}\label{fig:classification}
\end{figure}

\subsection{Performance Analysis of TET-GAN}

In addition to model benchmarking, we conducted experiments to further analyze the performance of TET-GAN.

\textbf{Ablation Study}.
In Fig.~\ref{fig:ablation1}, we study the effect of the reconstruction loss (Eq.~(\ref{eq:desty_feat_loss})) and the feature loss (Eq.~(\ref{eq:ae_rec_loss})). Without these two losses, even the color palette of the example style is not correctly transferred. In Fig.~\ref{fig:ablation1}(c), the glyph is not fully disentangled from the style, leading to bleeding artifacts. The satisfying results in Fig.~\ref{fig:ablation1}(d) verify that our feature loss effectively guides TET-GAN to extract valid content representations to synthesize clean text effects. For one-reference text effect transfer, as shown in Fig.~\ref{fig:ablation3}, if trained from scratch (namely, the self-supervised adversarial training scheme~\cite{Yang2018nonstationary}), the performance of our network drops dramatically, verifying that pretraining on our dataset successfully teaches our network the domain knowledge of text effect synthesis.

\textbf{Disentanglement of style and content features}.
TET-GAN disentangles content and style to enable multi-style transfer and removal.
To measure the entanglement of the two entities, we design experiments to analyze the representation capabilities of the extracted features.
Based on the intuition that a meaningful style (content) feature contains enough information to represent its style (content)
but little information to represent its content (style), we perform classification over both features.
Specifically, we extract the content feature and style feature of a style image as the output features of the sharing layers of $G_{\mathcal{X}}$ and $G_{\mathcal{Y}}$, respectively.
We additionally train an autoencoder over style images to extract the feature that preserves both the glyph and style information, which we denote as the AE feature for comparison.
Then, we train a five-layer classification network to predict the content and style label based on the input content feature, style feature and AE feature. Another six-layer classification network fed with the original style image is also trained as a reference. We elaborate the network architecture to make the two classification networks have similar numbers of parameters. The prediction accuracies during training over TE141K-C are plotted in Fig.~\ref{fig:classification}.

The AE feature is comparable to the image input. Compared with other inputs, the content feature extracted by TET-GAN contains highly representative and discriminative glyph information but very little style information. The same is true for the style feature, verifying the disentanglement of content and style.
This is because, for the content feature, $E^c_{\mathcal{Y}}$ is tasked with approaching the ideal glyph feature via our feature loss (Eq.~(\ref{eq:desty_feat_loss})). For the style feature, $E^s_{\mathcal{Y}}$ aims to extract style features that apply to arbitrary glyphs in content images, and is thus driven to eliminate the irrelevant glyph information from the style image.
In addition to the drive of the loss functions, another reason may be the good nature of our collected dataset, that is, high quality paired data with style labels.

\begin{figure}[t]
  \centering
  \includegraphics[width=0.98\linewidth]{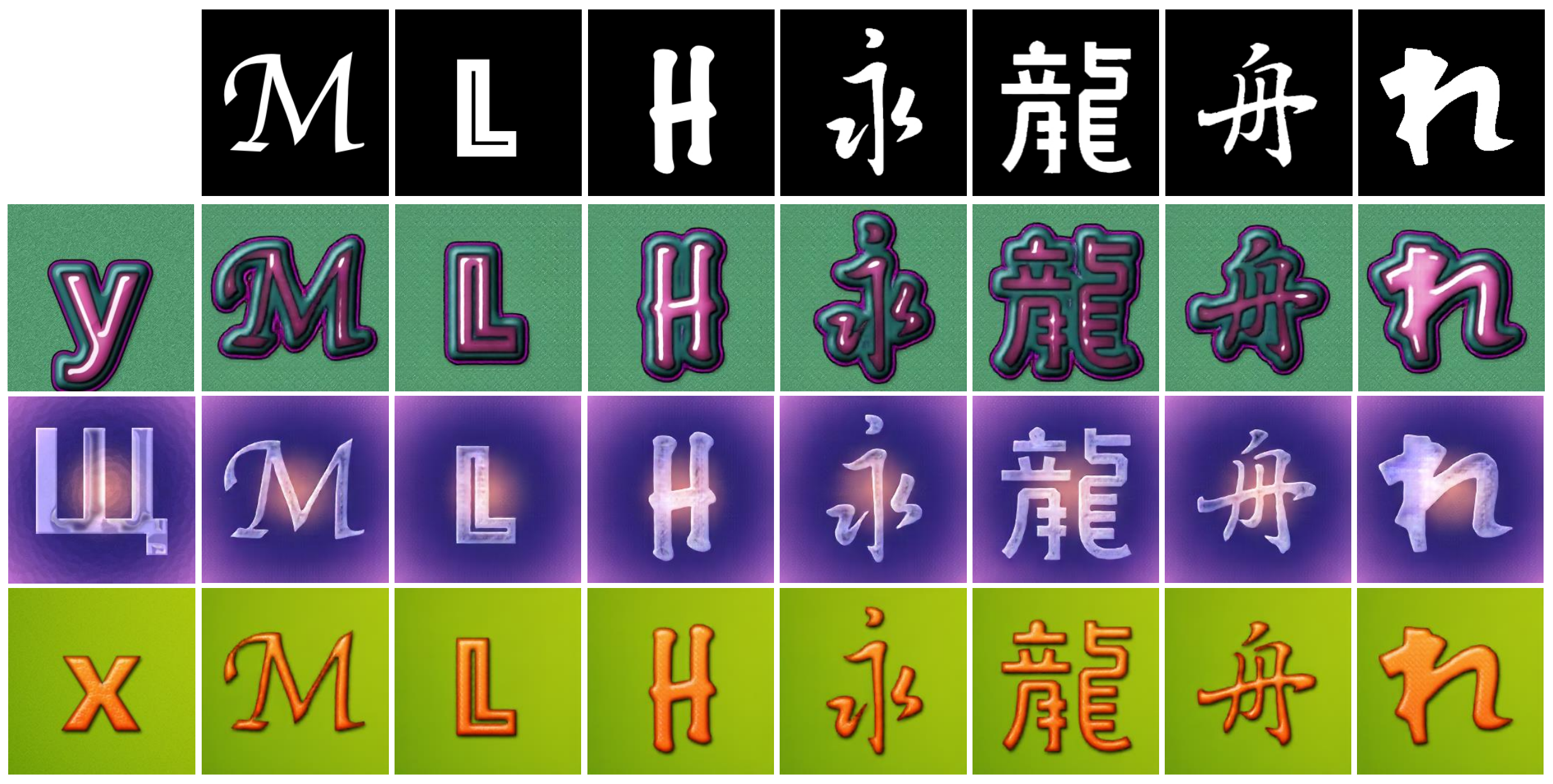}
  \caption{Stylization results in different font styles. First row: input text images. Next rows: input style images and the style transfer results.}\label{fig:font-robust}
\end{figure}

\textbf{Generalizability across font}.
To study the generalizability of TET-GAN across font styles, we select several text images in quite different fonts, as shown in Fig.~\ref{fig:font-robust}. Many of them are handwriting styles with irregular strokes.
TET-GAN successfully renders plausible text effects on these challenging strokes,
verifying its capability of transferring the text effects on a reference image to other glyphs in different font styles.

\begin{figure}[t]
  \centering
  \subfigure[Transferring one font style onto different characters]{
  \includegraphics[width=0.96\linewidth]{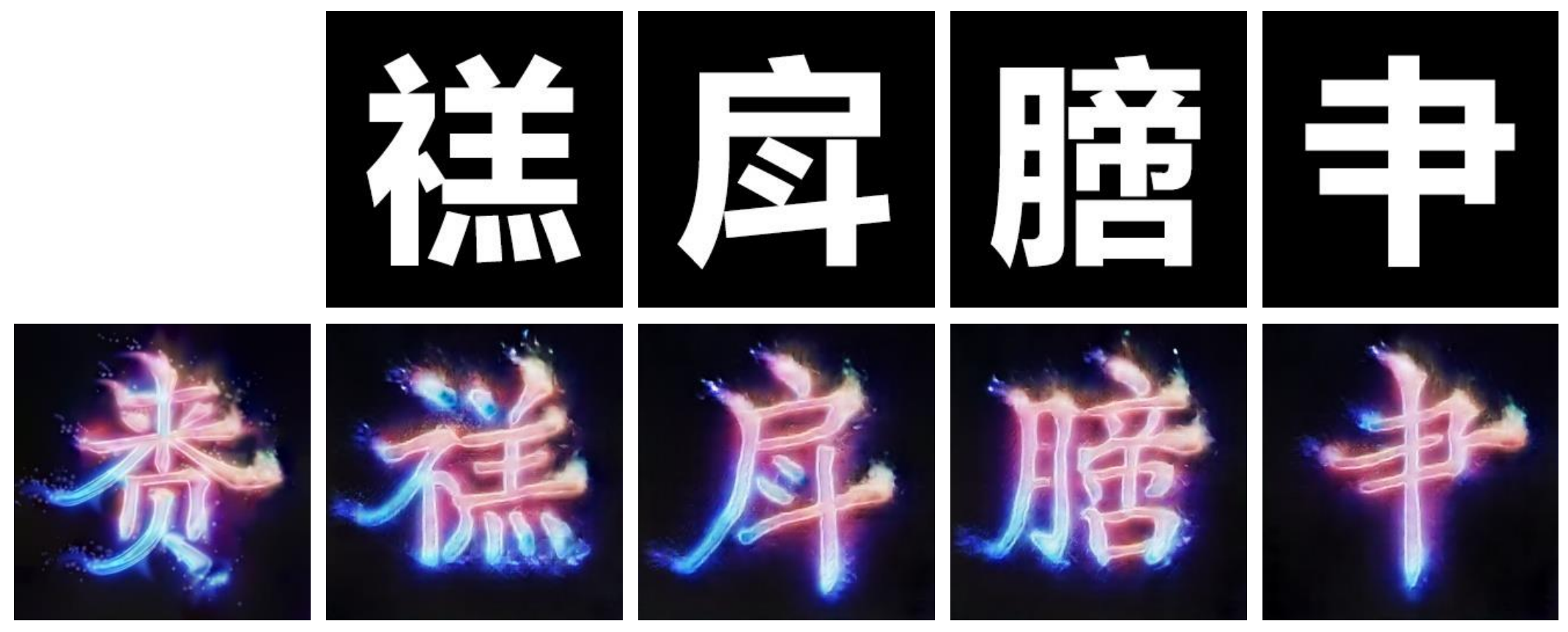}}
  \subfigure[Transferring different font styles onto one character]{
  \includegraphics[width=0.96\linewidth]{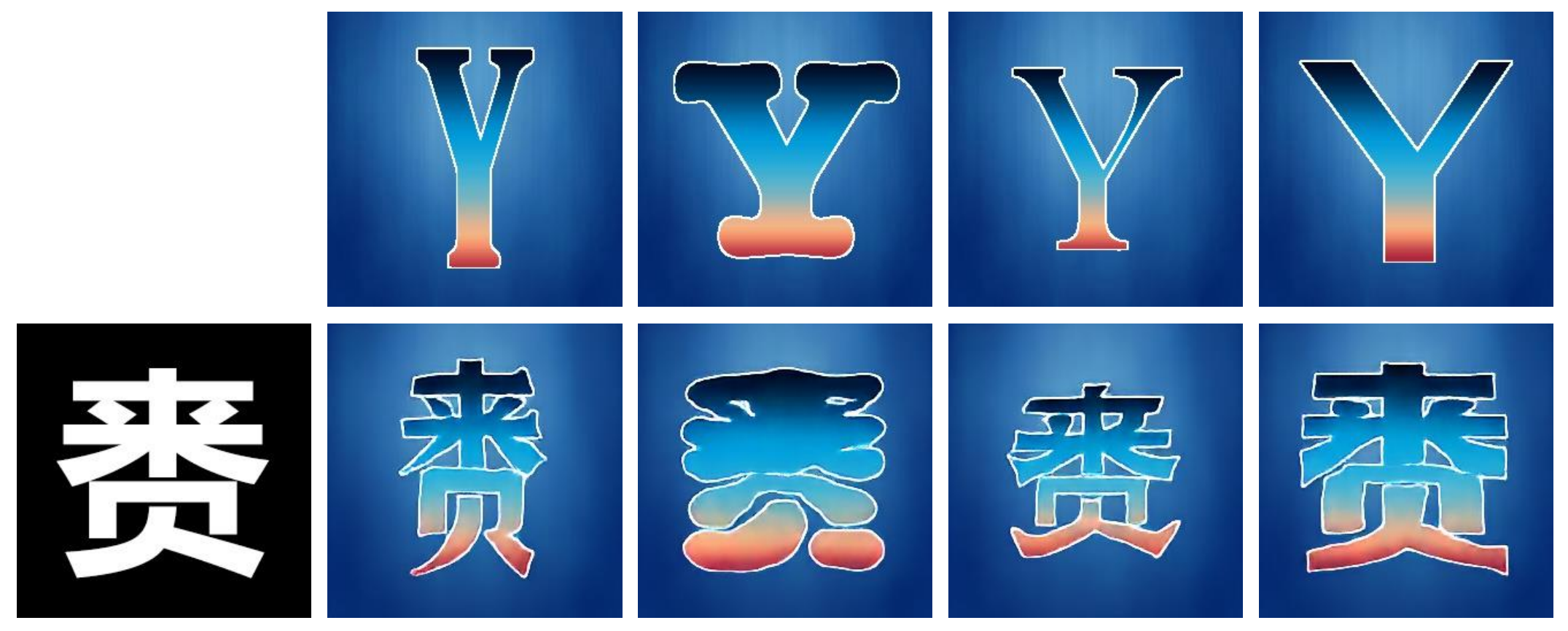}}
  \caption{Joint font style and text effect transfer results. (a) First row: input text images. Second row: input style image and the style transfer results. (b) First row: input style images. Second row: input text image and the style transfer results.}\label{fig:font-transfer}
\end{figure}

\textbf{Joint font style and text effect transfer}.
In Fig.~\ref{fig:font-transfer}, we study the performance of TET-GAN in joint font style and text effect transfer. We prepare a font dataset with $775$ Chinese characters in $30$ different font styles and train TET-GAN with the first $740$ characters. The style transfer results over the remaining unseen characters are shown in Fig.~\ref{fig:font-transfer}(a). It can be seen that the font style and text effects are effectively transferred.
In Fig.~\ref{fig:font-transfer}(b), we use English letters in TE141K-E as a reference style.
Note that both the glyph and the font styles are unseen during the training of our font style transfer model. TET-GAN selects the best-matched font among the $30$ training fonts based on the reference image to render the target text, which achieves satisfactory style consistency with the input.

\begin{figure}[t]
  \centering
  \includegraphics[width=0.98\linewidth]{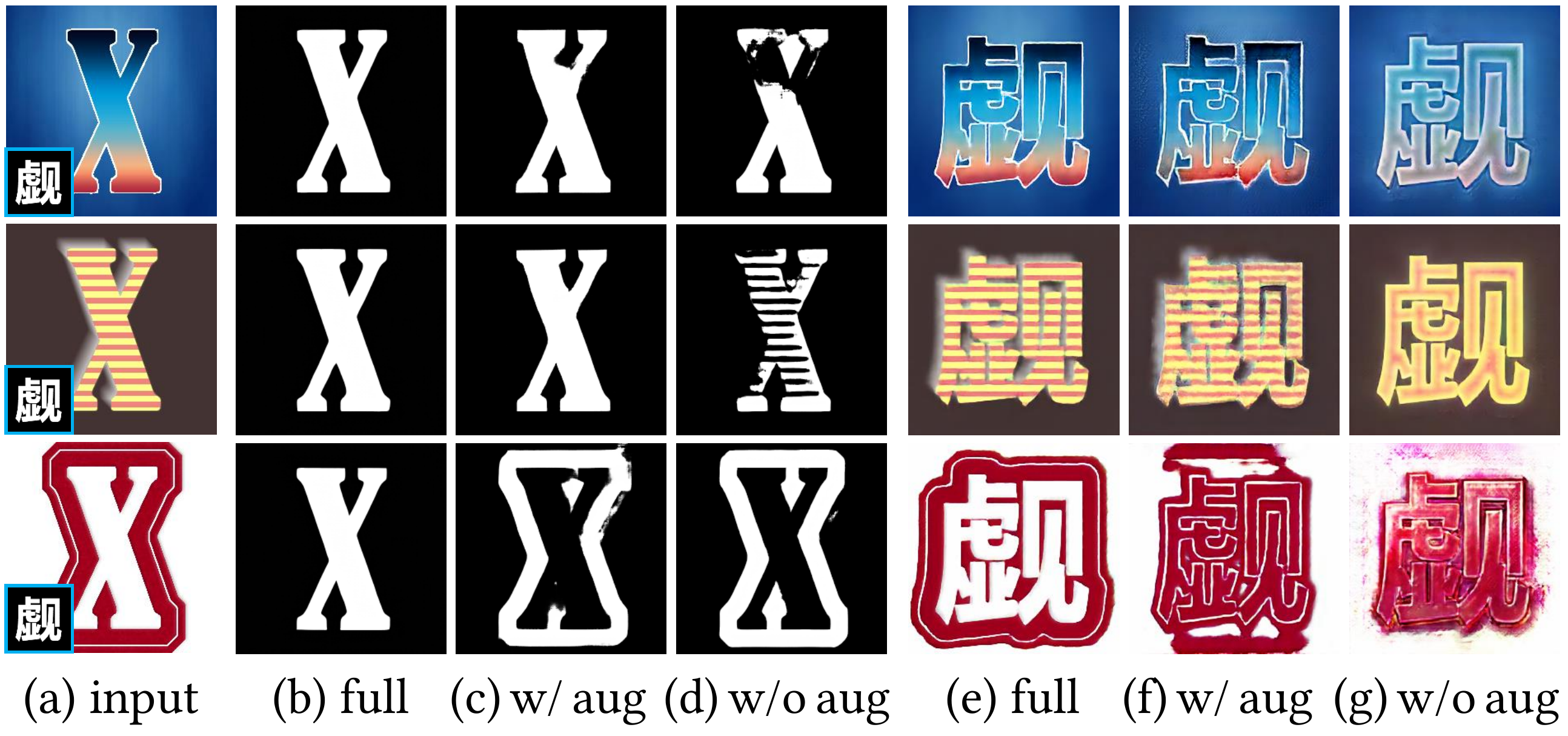}
  \caption{Destylization and stylization results of TET-GAN w/ and w/o adversarial augmentation. (a): Input example text effects from TE141K-E with the target text in the lower-left corner. (b)-(g): Destylization and stylization results by the models trained with supervision on the full TE141K (full), trained with supervision on TE141K-C and adversarially augmented with TE141K-E (w/ aug) and trained with only supervision on TE141K-C (w/o aug), respectively.}\label{fig:sei-results}
\end{figure}

\textbf{Semisupervised text effect transfer}.
In Fig.~\ref{fig:sei-results}, we compare the performance of supervised text transfer with and without adversarial augmentation. We manually divide TE141K-E into two parts with no overlap in glyphs to generate our unpaired data. The model trained only on TE141K-C serves as a baseline to show the performance of our model on unseen glyphs and styles. Meanwhile, the model trained on the full TE141K gives an upper bound that we can expect from unsupervised learning.
It can be seen that our semisupervised model learns from unpaired data to better imitate the target styles. We also observe a clear improvement in the destylization of English glyphs that is fairly different from that of the training data in TE141K-C, indicating better glyph generalization of the model.
However, our semisupervised model fails to handle the challenging text effects in the third row of Fig.~\ref{fig:sei-results},
verifying that our dataset is challenging and there is still much room for improvement in semisupervised learning.

\begin{figure}[t]
  \centering
  \includegraphics[width=0.96\linewidth]{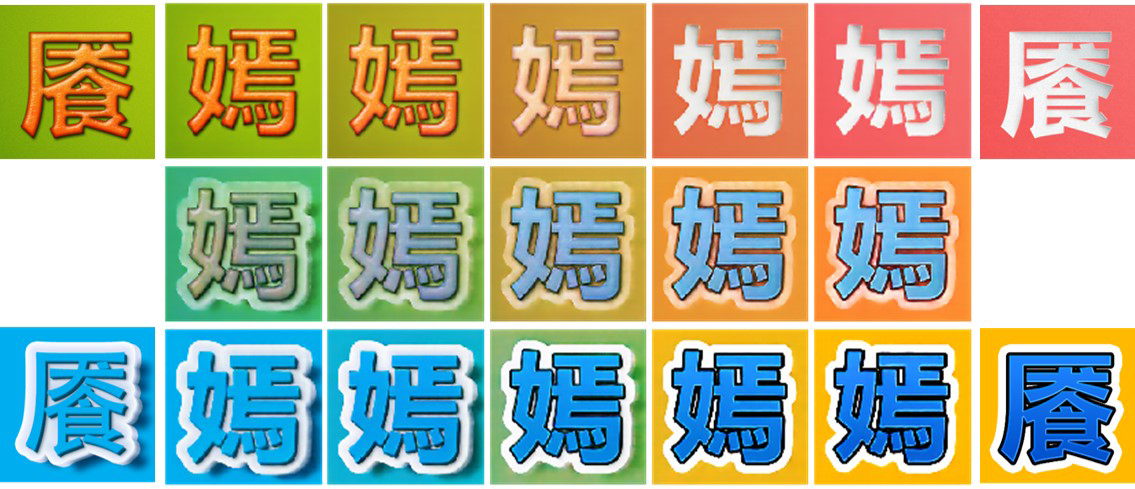}
  \caption{Applications of TET-GAN for style interpolation.}\label{fig:app}
\end{figure}

\textbf{Style interpolation}. The flexibility of TET-GAN is further shown by the application of style interpolation.
The explicit style representations enable intelligent style editing. Fig.~\ref{fig:app} shows an example of style fusion. We interpolate between four different style features and decode the integrated features back to the image space, obtaining new text effects.

\begin{figure}[t]
  \centering
  \includegraphics[width=0.98\linewidth]{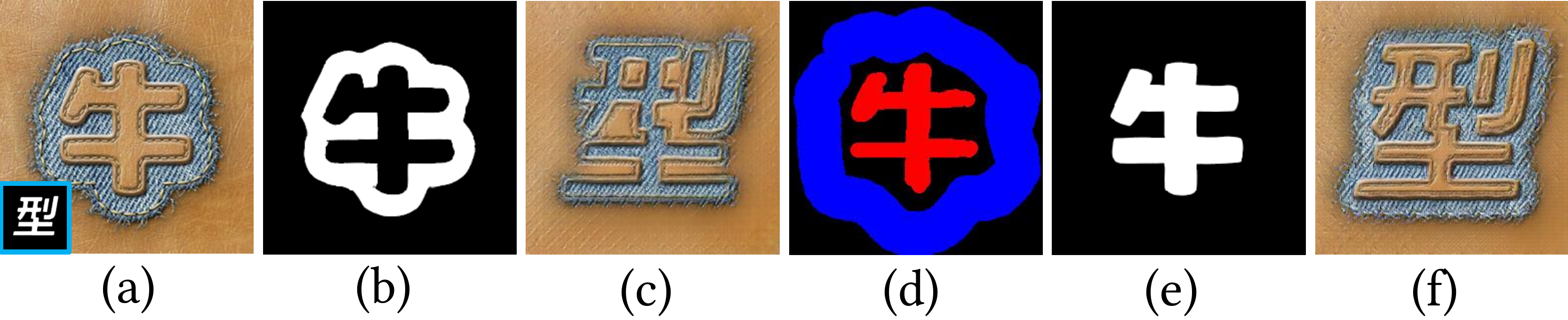}
  \caption{User-interactive unsupervised style transfer. (a): Example text effects and the target text. (b),(c): Our destylization and stylization results after finetuning. (d): A mask provided by the user, where the blue and red regions indicate the background and foreground, respectively. (e),(f): Our destylization and stylization results with the help of the mask.  }\label{fig:fail}
\end{figure}

\textbf{Failure case and user interaction}. While our approach has generated appealing results, some limitations still exist.
Our destylization subnetwork is not fool-proof due to the extreme diversity of the text effects, which may differ completely from our collected text effects.
Fig.~\ref{fig:fail} shows a failure case of one-reference unsupervised text effect transfer.
Our network fails to recognize the glyph. As a result, in the stylization result, the text effects in the foreground and background are reversed.
This problem can possibly be solved by user interaction. Users can simply paint a few strokes (Fig.~\ref{fig:fail}(d)) to provide a priori information about the foreground and the background, which is then fed into the network as guidance to constrain glyph extraction.
Specifically, let $M_f$ and $M_g$ be the binary mask of foreground and background (\textit{i.e.}, the red channel and the blue channel in Fig.~\ref{fig:fail}(d)) provided by the user, respectively.
Then, a guidance loss is added to Eq.~(\ref{eq:total_loss}):
\begin{equation}
\begin{aligned}
  \mathcal{L}_{\text{guid}}=&~\mathbb{E}_{y}[\|G_{\mathcal{X}}(E^{c}_{\mathcal{Y}}(y))\odot M_f-M_f\|_{1}]\\
  +&~\mathbb{E}_{y}[\|G_{\mathcal{X}}(E^{c}_{\mathcal{Y}}(y))\odot M_b-\mathbf{0}\|_{1}],
\end{aligned}
\end{equation}
where $\odot$ is the elementwise multiplication operator.
As shown in Fig.~\ref{fig:fail}(e), under the guidance of $\mathcal{L}_{\text{guid}}$, the glyph is correctly extracted, and the quality of the style transfer result (Fig.~\ref{fig:fail}(f)) is thereby greatly improved.

\section{Conclusion}
\label{sec:conclusion}

In this paper, we introduce a novel text effects dataset with 141K text effect/glyph pairs in total, which consists of 152 professionally designed text effects and 3K different kinds of glyphs, including English letters, Chinese characters, Japanese kana, and Arabic numerals.
Statistics and experimental results validate the challenges of the proposed dataset. In addition, we design effective data preprocessing and augmentation methods that can improve the robustness of transfer models. Moreover, we present a novel TET-GAN for text effect transfer. We integrate stylization and destylization into one uniform framework to jointly learn valid content and style representations of the artistic text. The benchmarking results demonstrate the superiority of TET-GAN in generating high-quality artistic typography. As a future direction, one may explore other more sophisticated style editing methods on the proposed dataset, such as background replacement, color adjustment and texture attribute editing. We believe the proposed dataset has the potential to boost the development of corresponding research areas.

\bibliographystyle{IEEEtran}
\bibliography{bibliography}

\ifCLASSOPTIONcaptionsoff
  \newpage
\fi

\end{document}